\pdfoutput=1
\documentclass[10pt,twocolumn,letterpaper]{article}

\usepackage{iccv}
\usepackage{times}
\usepackage{epsfig}
\usepackage{graphicx}
\usepackage{amsmath}
\usepackage{amssymb}

\newcommand{\benchname}{\textsc{EqBen}\xspace}
\newcommand{\algname}{\textsc{EqSim}\xspace}
\newcommand{\algnamev}{\textsc{EqSim}$_{\textrm{v1}}$\xspace}
\newcommand{\algnamevv}{\textsc{EqSim}$_{\textrm{v2}}$\xspace}

\newtheorem{definition}{Definition}

\usepackage{wasysym}
\usepackage{makecell}

\usepackage{bbding}
\usepackage{caption}
\usepackage{enumitem}

\usepackage{wrapfig}
\usepackage{multirow}
\usepackage{booktabs}
\usepackage[table,xcdraw]{xcolor}
\usepackage{xcolor}         
\usepackage{arydshln}

\definecolor{mygray}{gray}{0.9}

\usepackage{wrapfig}

\definecolor{LightCyan}{rgb}{0.9059,0.9961,1}
\usepackage{multirow}
\usepackage{graphicx}
\usepackage[font=small,labelfont=bf,aboveskip=1pt,belowskip=-8pt]{caption}  
\usepackage{makecell}
\usepackage{tabulary}
\definecolor{demphcolor}{RGB}{144,144,144}

\usepackage{pifont}
\newcommand{\inlineimg}[1]{\raisebox{-0.2\baselineskip}{\includegraphics[height=0.95\baselineskip]{#1.png}}}

\newlength\savewidth\newcommand\shline{\noalign{\global\savewidth\arrayrulewidth
  \global\arrayrulewidth 1pt}\hline\noalign{\global\arrayrulewidth\savewidth}}

\newcommand{\tablestyle}[2]{\setlength{\tabcolsep}{#1}\renewcommand{\arraystretch}{#2}\centering\footnotesize}
\makeatletter\renewcommand\paragraph{\@startsection{paragraph}{4}{\z@}
  {.5em \@plus1ex \@minus.2ex}{-.5em}{\normalfont\normalsize\bfseries}}\makeatother
\newdimen\abovecrulesep
\newdimen\belowcrulesep

\newcommand{\eat}[1]{}

\usepackage[pagebackref=true,breaklinks=true,letterpaper=true,colorlinks,bookmarks=false]{hyperref}

\usepackage[capitalize]{cleveref}
\crefname{section}{Sec.}{Secs.}
\Crefname{section}{Section}{Sections}
\Crefname{table}{Table}{Tables}
\crefname{table}{Tab.}{Tabs.}

\iccvfinalcopy 

\def\iccvPaperID{3736} 

\begin{document}

\title{Equivariant Similarity for Vision-Language Foundation Models}

\author{Tan Wang$^{1}$, 
Kevin Lin$^{2}$, 
Linjie Li$^{2}$,
Chung-Ching Lin$^{2}$,
Zhengyuan Yang$^{2}$,\\
Hanwang Zhang$^{1}$,
Zicheng Liu$^{2}$,
Lijuan Wang$^{2}$\\
\normalsize$^1$Nanyang Technological University~~
$^2$Microsoft\\
\footnotesize {\texttt{\{TAN317,hanwangzhang\}@ntu.edu.sg}},~
\footnotesize {\texttt{\{keli,lindsey.li,chungching.lin,zhengyang,zliu,lijuanw\}@microsoft.com}}
\\
}

\maketitle


\begin{abstract}
This study explores the concept of equivariance in vision-language foundation models (VLMs), focusing specifically on the multimodal similarity function that is not only the major training objective but also the core delivery to support downstream tasks. 
Unlike the existing image-text similarity objective which only categorizes matched pairs as similar and unmatched pairs as dissimilar,  
equivariance also requires similarity to vary faithfully according to the semantic changes. 
This allows VLMs to generalize better to nuanced and unseen multimodal compositions. 
However, modeling equivariance is challenging as the ground truth of semantic change is difficult to collect.
For example, given an image-text pair about a dog, it is unclear to what extent the similarity changes when the pixel is changed from \texttt{dog} to \texttt{cat}? 
To this end, we propose \algname, a regularization loss that can be efficiently calculated from any two matched training pairs and easily pluggable into existing image-text retrieval fine-tuning.
Meanwhile, to further diagnose the equivariance of VLMs, we present a new challenging benchmark \benchname. Compared to the existing evaluation sets, \benchname is the first to focus on ``visual-minimal change''. 
Extensive experiments show the lack of equivariance in current VLMs\footnote{We also include results of Multimodal LLM in Section~\ref{sec:mllm}.} and validate the effectiveness of \algname\footnote{Code is available at \url{https://github.com/Wangt-CN/EqBen}}.

\end{abstract}

\vspace{-0.1in}
\section{Introduction}
Vision-language (VL) training is all about learning ``good'' features for each modality, 
such that the features should faithfully represent the underlying semantics. Thanks to the large-scale image-text pairs on the Web, we have abundant multimodal supervision for the two features with the same semantic meaning~\cite{tan-bansal-2019-lxmert,li2021align,radford2021learning,jia2021scaling}---each matched image-text pair should have ``similar'' visual and textual features, and each unmatched pair should have ``dissimilar'' ones. Thus, the image-text similarity plays a crucial role to define the feature quality in training VL foundation models (VLMs)~\cite{wang2022image,yu2022coca,li2021align,radford2021learning,jia2021scaling,dou2021empirical,dou2022coarse,li2022blip}.

\begin{figure}[t]
    \centering
    \footnotesize
    \includegraphics[width=.48\textwidth]{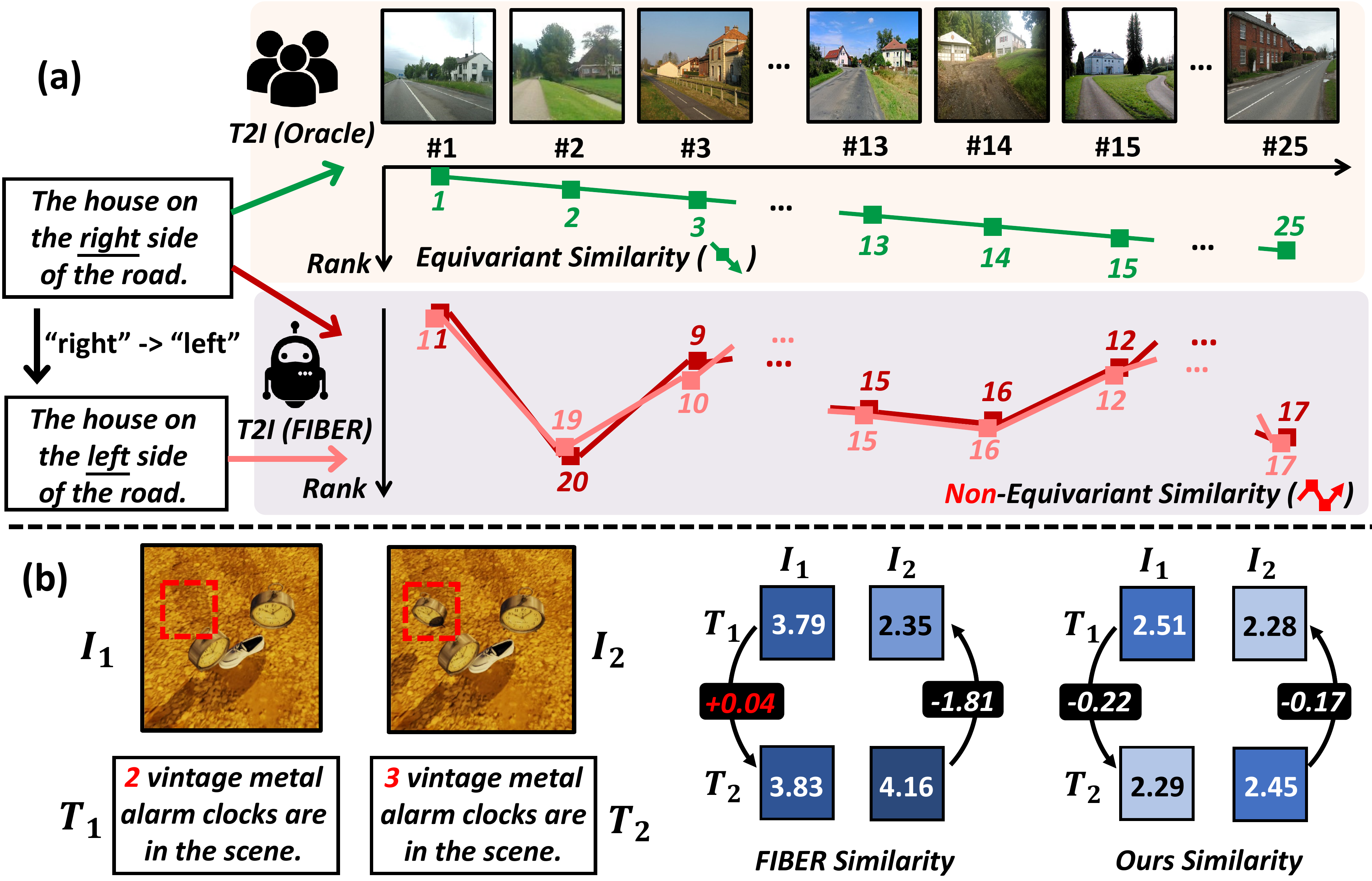}
    \caption{(a) 
    Comparison between oracle and the latest SoTA VLM FIBER~\cite{dou2022coarse} similarity measure by ranking the candidate images with the given query texts. Check Appendix for full ranking results. (b) Measuring the similarity score change (number in $\blacksquare$) of FIBER~\cite{dou2022coarse} and our proposed \algname by applying a slight text change (``2''$\leftrightarrow$``3''). Darker color indicates larger similarity. 
    }
    \vspace{-4mm}
    \label{fig:intro_fig}
\end{figure}

Has the prevailing ``matched \textit{vs.} unmatched'' similarity fulfilled its duty? Yes and no. On the one hand, recent VLMs~\cite{rombach2022high,wang2022image,dou2022coarse,radford2021learning,yu2022coca,ramesh2022hierarchical} have demonstrated impressive results in various downstream VL tasks such as image-text retrieval.   
However, on the other hand, it is acknowledged by the community that the VLMs still fall short in \textit{nuanced and complex semantic compositions}~\cite{ramesh2022hierarchical,cho2022dall,parcalabescu2021valse,thrush2022winoground}.
In this regard, we present a text-to-image retrieval example on LAION400M~\cite{schuhmann2021laion} with the most recent SOTA VLM FIBER~\cite{dou2022coarse}.
As shown in Figure~\ref{fig:intro_fig}(a), given the query text ``the house on the \emph{right} side of the road'', we first invite 5 graduate students to rank 25 candidate images from most similar to least similar. The continuously decreasing ranking from human judges (\inlineimg{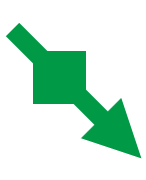}) is served as the oracle semantic similarity measure.
We then compared this ranking with the ones from FIBER~\cite{dou2022coarse} (\inlineimg{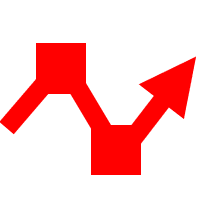}). 
Although FIBER correctly retrieved the top-$1$ image (image\#1, ranks 1), some semantically incorrect images (\eg, image\#25, ranks 17) are falsely ranked higher than the correct ones (\eg, image\#2, ranks 20).
Furthermore, when modifying the query text with a slight semantic change (``\underline{right}'' $\rightarrow$ ``\underline{left}''), the rankings remain almost the same.
Clearly, the similarity changes in FIBER do not faithfully reflect the semantic changes in images (\#1 $\rightarrow$ \#25) or text queries (``\underline{right}'' $\rightarrow$ ``\underline{left}'').

To quantitatively measure the above inconsistency between semantic and similarity score changes, we consider two matched image-text pairs $\{I_1, T_1\}$ and $\{I_2, T_2\}$ that are semantically similar but only different in the number of clocks in Figure~\ref{fig:intro_fig} (b).  
With a slight change of clock counts in caption (``2''$\rightarrow$``3''), FIBER mistakenly assigns a higher similarity score to $\{I_1,T_2\}$ rather than $\{I_1,T_1\}$ ($3.83$ v.s. $3.79$). Furthermore, the changes in similarity scores guided by the semantic change (``2''$\leftrightarrow$``3'')
are highly inconsistent ($+0.04$ v.s. $-1.81$). 
Ideally, an \textbf{\emph{equivariant}} image-text similarity measure should faithfully reflect the semantic change, \ie, the same semantic changes should lead to a similar amount of similarity changes (\eg, $-0.22$ v.s. $-0.17$ of ours in Figure~\ref{fig:intro_fig}(b)).

\begin{figure}[t]
    \centering
    \footnotesize
    \includegraphics[width=.47\textwidth]{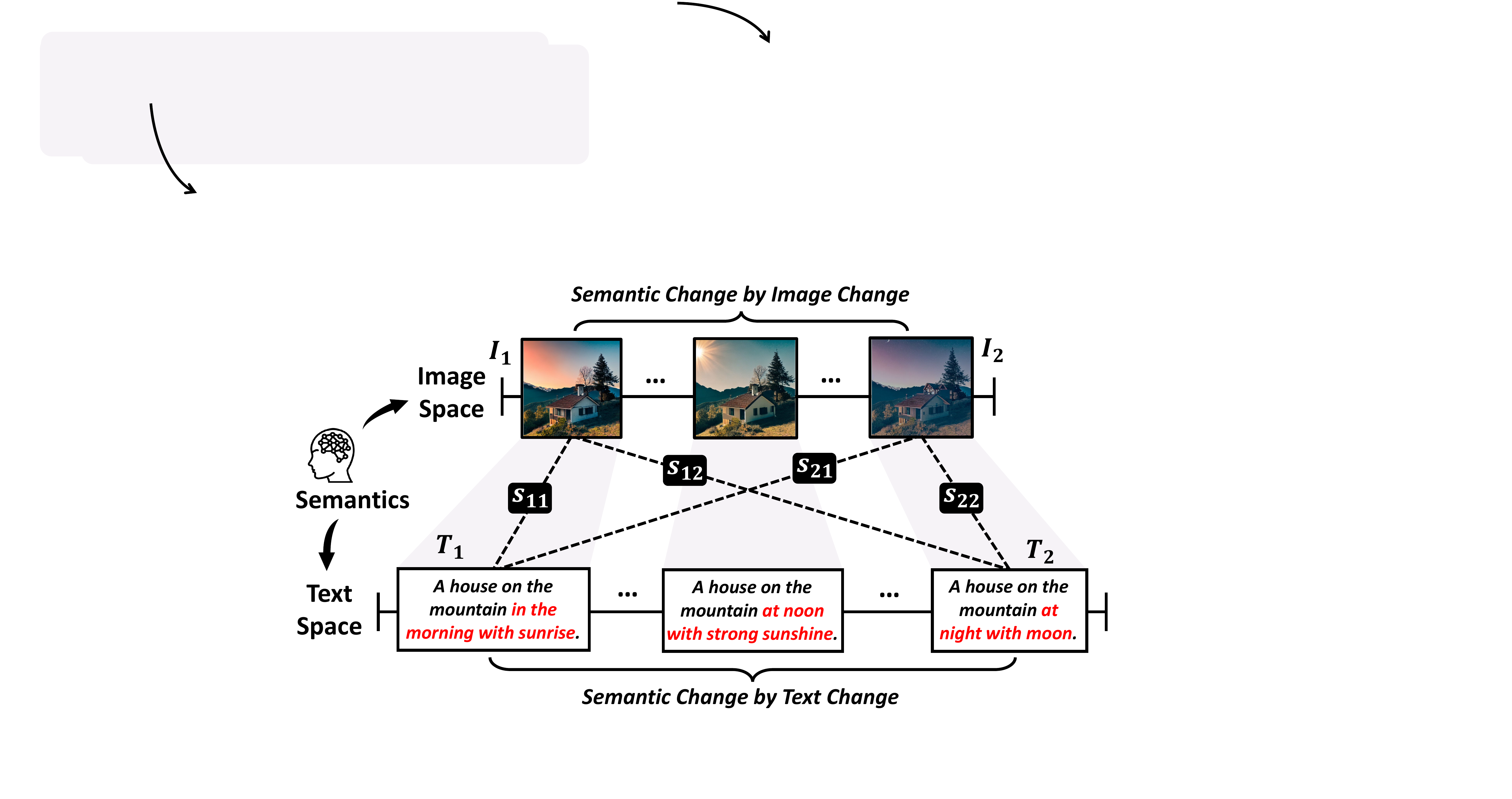}
    \caption{The illustration of the core idea in \algname. Besides the two matched pairs $\{I_1, T_1\}$ and $\{I_2, T_2\}$, we don't need extra annotation such as the middle pair. 
    }
    \vspace{-3mm}
    \label{fig:intro_semantic}
\end{figure}

\noindent\textbf{Equivariance Loss}. To address this non-equivariance issue, we propose Equivariant Similarity Learning (\algname), which imposes additional equivariance regularization on image-text pairs for VLM learning without additional supervision. Figure~\ref{fig:intro_semantic} illustrates the underlying semantics perceived by human, where each matched pair demonstrates the image and text corresponding to the underlying semantic. 
Given two matched image-text pairs $\{I_1,T_1\}$ as semantic 1 and $\{I_2,T_2\}$ as semantic 2, we can obtain four similarity scores $s_{11}$, $s_{12}$, $s_{22}$, and $s_{21}$. We define \textbf{\emph{Equivariant Similarity}} to be an image-text similarity function, whose output value should correspond to the underlying semantic change, which can be measured by text or image change. 

\vspace{-4pt}
\begin{definition} (Equivariant Similarity)
The similarity $s$ between image and text is equivariant if and only if the following equations hold:

\vspace{-4mm}
{\small{
\begin{equation}
s_{11}-s_{12} =  \underbrace{\sum\nolimits_{T_1}^{T_2} \mu(T)},
~~~s_{22}-s_{21} =  \underbrace{\sum\nolimits_{T_2}^{T_1} \mu(T)},
\vspace{-0.2cm}
\label{eq:intro_eq1}
\end{equation}
}
\hspace{20mm} Semantic Change Measured by Text Change
}

\vspace{-0.4cm}
{\small{
\begin{equation}
s_{11}-s_{21} =  \underbrace{\sum\nolimits_{I_1}^{I_2} \mu(I)},
~~~s_{22}-s_{12} =  \underbrace{\sum\nolimits_{I_2}^{I_1} \mu(I)},
\vspace{-0.2cm}
\label{eq:intro_eq2}
\end{equation}
\hspace{20mm} Semantic Change Measured by Image Change
}}\label{def:eqsim}
\end{definition}
where $\mu(I)$ ($\mu(T)$) denotes the measure~\cite{royden1988real} in image (text) space, \ie, an infinitesimal unit of visual (textual) change. 
Based on Definition~\ref{def:eqsim}, we formally derive \algname, an equivariance loss for a hybrid learning strategy on both semantically close and distant training pairs (Section~\ref{sec:method}). Specifically, \algname directly enforces $s_{11}-s_{12}=s_{22}-s_{21}$ and $s_{11}-s_{21}=s_{22}-s_{12}$ for semantically close samples; while for semantically distant samples, we derive a simplified formulation of $s_{12}=s_{21}$.
We show that adding \algname as a regularization term improves existing similarity training objectives significantly on challenging datasets (\eg, over $4\%$ on Winoground~\cite{thrush2022winoground}) and tricky tasks (\eg, around $30\%$ on VALSE~\cite{parcalabescu2021valse}). \algname can also retain or even improve retrieval performance on Flickr30K~\cite{plummer2015flickr30k} dataset.

\noindent\textbf{Equivariance Benchmark}. To further facilitate the proper evaluation of equivariance in VL community, we present a novel evaluation benchmark dubbed \benchname (Section \ref{sec:dataset}). 
Motivated by the examples in Figure~\ref{fig:intro_fig}(b), \benchname features ``slightly'' mis-matched pairs with a \emph{minimal semantic drift} from the matched pairs, as opposed to ``very different'' matched and unmatched pairs that are easily distinguishable by both non-equivariant and equivariant similarities. 
Unlike recent efforts~\cite{parcalabescu2021valse,thrush2022winoground} focusing on minimal semantic changes in captions, \benchname pivots on diverse \textit{visual}-minimal changes, automatically curated from time-varying visual contents in natural videos and synthetic engines with more precise control. 
We benchmark a full spectrum of VLMs on \benchname, and reveal that the non-equivariant similarity in existing VLMs fails easily. On this new test bed, \algname can serve as a remedy and bring a large performance gain of $\sim$3\% on average.

Our contributions are summarized as follows: \textbf{(1)} We comprehensively study the problem of similarity equivariance in VLMs. We propose \algname for equivariant training and \benchname for diagnostic evaluation; \textbf{(2)} \algname is not only theoretically grounded but also simple, effective and easily pluggable; and \textbf{(3)} \benchname clearly diagnoses that conventional evaluation is not responsive to equivariance. Furthermore, \algname can significantly improve VLMs on \benchname, as well as other challenging benchmarks.

\section{Related Work}

\noindent\textbf{Pre-training VL Models}.
Early object detector (OD)-based methods~\cite{chen2020uniter, zhang2021vinvl, li2020oscar, li2019visualbert, lu2019vilbert, gan2020large, tan-bansal-2019-lxmert} utilized the offline image region features from a pre-trained object detector~\cite{ren2015faster}. 
More recent methods mainly learn from image pixels directly in an end-to-end  manner~\cite{wang2021simvlm, jia2021scaling, wang2021vlmo, singh2021flava, li2022blip, wang2022unifying, yu2022coca, alayrac2022flamingo}.
Researchers~\cite{gan2022vision} further categorize VLMs into (1) Dual-Encoder (\eg, CLIP~\cite{radford2021learning} and ALIGN~\cite{li2021align}) and (2) Fusion-Encoder (\eg, METER~\cite{dou2021empirical}, FIBER~\cite{dou2022coarse}, and ALBEF~\cite{li2021align}).
It is worth noting that our proposed \algname is model-agnostic, and can be easily plugged into the image-text alignment objectives such as Image-Text Matching (ITM) and Image-Text Contrastive (ITC) loss.

\noindent\textbf{Diagnosing VL Models}.
Years of VL research have spawned a series of VL evaluation kits, 
from classical VL tasks~\cite{zellers2019recognition, vinyals2015show,Yu2016ModelingCI,plummer2015flickr30k} (\eg, VQA~\cite{antol2015vqa} and image captioning~\cite{chen2015microsoft}), to more complex contexts, such as adversarial examples~\cite{li2021adversarial, chen2017attacking}, robustness~\cite{gupta2022grit,cadene2019rubi, wang2020visual, tang2020unbiased,jimenez2022carets} and counterfactual reasoning~\cite{shekhar2017foil,hendricks2021probing,hu2019evaluating,niu2021counterfactual}.
However, 
these benchmarks require manual annotation and their evaluation relies on task-specific 
model fine-tuning. 
Another line of work~\cite{thrush2022winoground,parcalabescu2021valse,zhao2022vl} probe VLMs on similarity measure
with minimal \emph{caption} semantic changes while keeping images intact. 
While our \benchname tries to test whether the inherent image-text similarity measure in existing VLMs is sensitive to visual semantic changes. 
The most relevant work is ImageCoDe~\cite{krojer-etal-2022-image} which leverages video frames toward fine-grained image-text retrieval. However, ImageCoDe requires additional human crowdsourcing and is limited to real-world video sources. 
In contrast, \benchname explores both natural and synthetic ways to generate image pairs with minimal semantic change, making the data generation process 
inclusive, automatic, and extensive.

\noindent\textbf{Equivariance Learning}.
Unlike the wide usage of invariance in deep neural networks (\eg, shift invariance achieved by convolutional layers), strict group equivariance~\cite{cohen2018spherical, cohen2016group, weiler2019general,bronstein2021geometric} is hard to apply in practice. 
However, the equivariance property still plays an important role in various fields, such as self-supervised learning~\cite{dangovski2021equivariant, xie2022should, wang2021self,patrick2020support,han2020self}, representation learning~\cite{qi2020learning}, and language understanding~\cite{gordon2019permutation}.
In this paper, we point out the significance 
of the equivariant similarity measure in VLMs. Based on this, we further propose a novel loss \algname for the regularization of equivariance, as well as a new challenging benchmark \benchname to diagnose the equivariance of existing VLMs. 
We notice that the recent CyCLIP~\cite{goel2022cyclip} delivers a similar idea but with different motivation, implementation and evaluation settings. In Table~\ref{tab:abla_arch}, we compare with CyCLIP-equivalent baseline as \algnamev. Check more detailed comparison in Appendix.

\section{Improving VLMs with \textsc{EqSim}}
\label{sec:method}

\begin{wrapfigure}{l}{0.13\textwidth}
\vspace{-4mm}
    \centering
    \footnotesize
    \includegraphics[width=.14\textwidth]{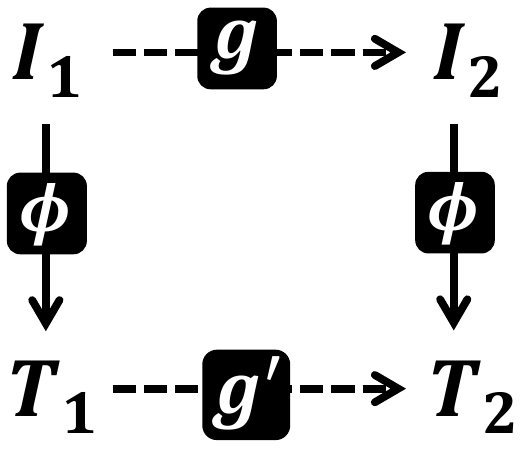}
    \caption{Commutativity in equivariant map.
    }
    \vspace{-2mm}
    \label{fig:eqmap}
\end{wrapfigure}
Recall that VLMs adopt the image-text similarity as the training proxy for learning multimodal feature representation~\cite{radford2021learning}. Therefore, the ultimate goal of pursuing equivariant similarity is to learn an equivariant feature map between image space and text space.

\begin{figure}[t]
    \centering
    \footnotesize
    \includegraphics[width=.45\textwidth]{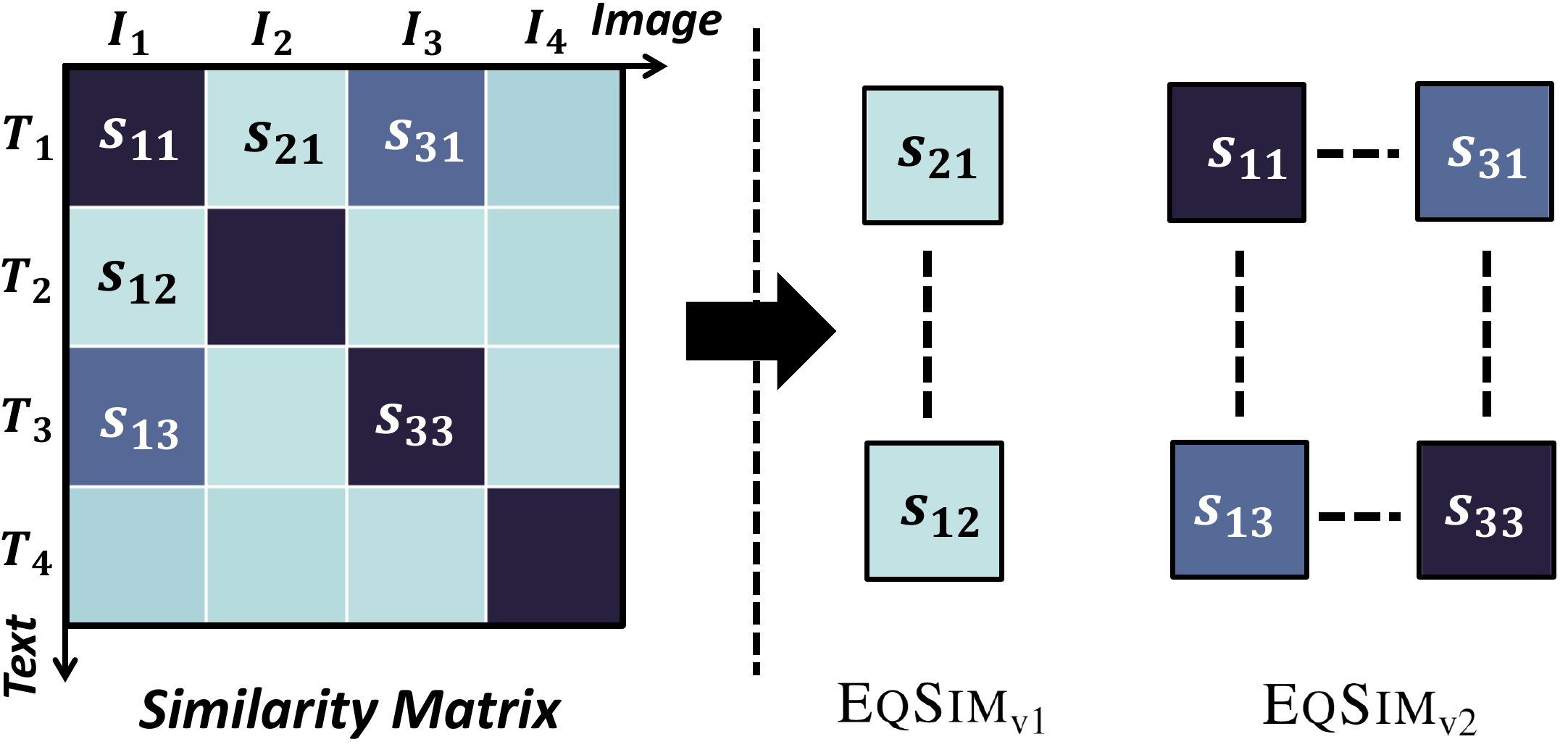}
    \caption{Illustration of \algname given a similarity matrix during training. 
    Darker color indicates higher similarity.
    }
    \label{fig:eqsim}
\end{figure}

\begin{figure*}[t]
    \centering
    \includegraphics[width=.98\textwidth]{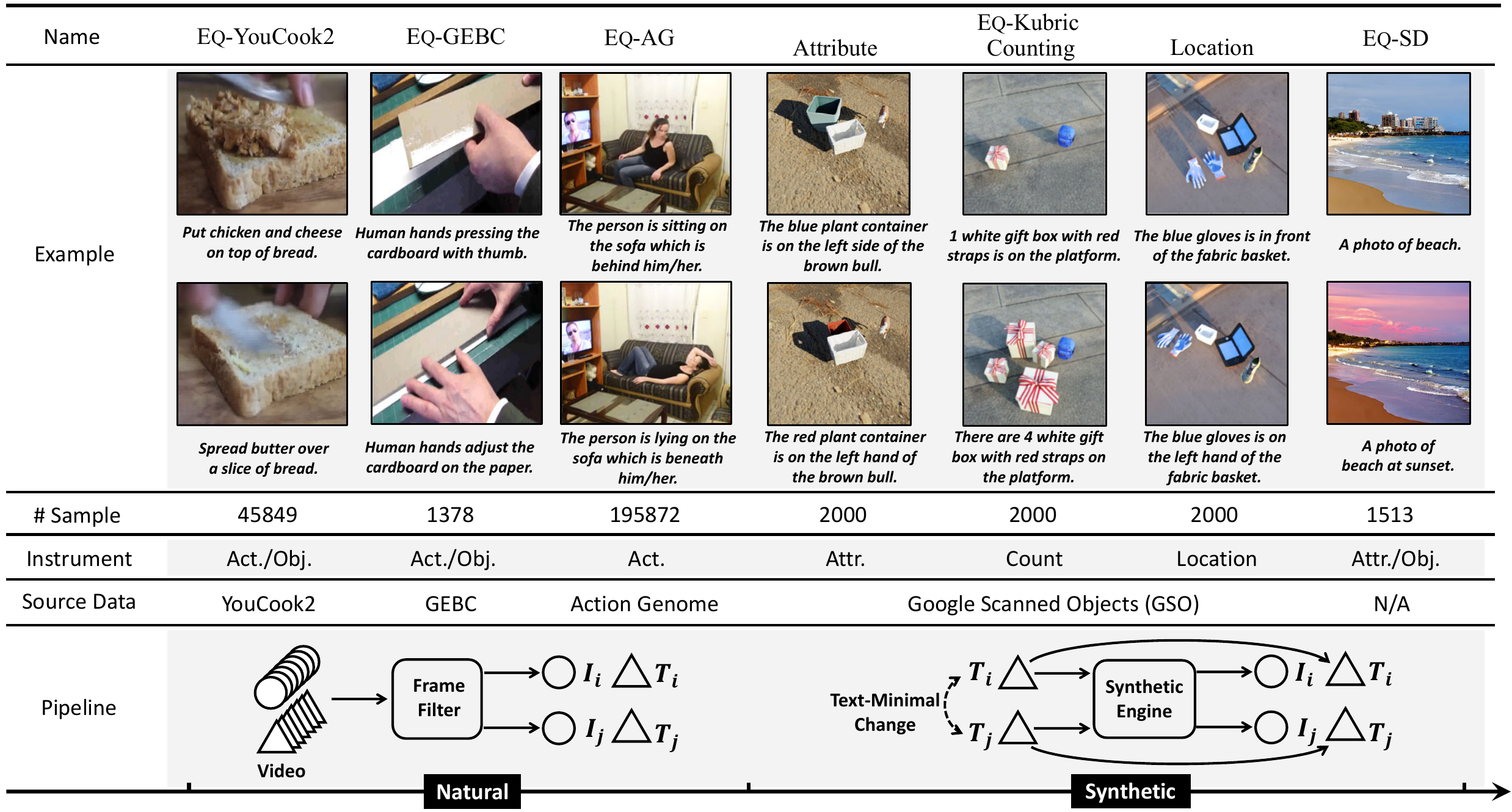}
    \caption{Overview of the proposed benchmark \benchname, which consists of 5 sub-datasets and can be categorized to natural and synthetic. 
    Act., Obj., Attr. denote action, object and attribute, respectively. 
    }
    \label{fig:benchmark}
\end{figure*} 

\begin{definition}
Let $\mathcal{I}$ and $\mathcal{T}$ be two continuous feature spaces. Let $\mathcal{G}$ be a group whose group action on $\mathcal{I}$ is defined by $g: \mathcal{I}\to \mathcal{I}$, and that on $\mathcal{T}$ is defined by $g': \mathcal{T}\to\mathcal{T}$. Then, $\phi: \mathcal{I}\to\mathcal{T}$ is an equivariant feature map if and only if $g'\cdot \phi(I) =\phi(g\cdot I)$ for all the group actions and $I\in\mathcal{I}$. The commutativity for $\phi$, $g$ and $g'$ is shown in Figure~\ref{fig:eqmap}.
\label{def:eqmap}
\end{definition}
In Definition~\ref{def:eqsim}, the measure $\mu$ can be considered as the semantic group acting on an infinitesimal region in image or text space. Thus, by applying the commutativity of Definition~\ref{def:eqmap} in Definition~\ref{def:eqsim} to change the sum from image space to text space. Without loss of generality, we only show the results of sum $1\to2$:
{\small{
\begin{equation}
\sum\nolimits_{T_1}^{T_2} \mu(T) = \sum\nolimits_{\phi(I_1)}^{\phi(I_2)} \mu(\phi(I)) = \sum\nolimits_{\phi(I_1)}^{\phi(I_2)} \phi(\mu(I)).
\end{equation}
\label{eq:method_eq3}
}}
\!\!\!This implies that the equivariant map establishes an isometry for the measure  $\mu(I)$ in image space and $\phi(\mu(I))$ in text space. Thus, they only differ by a constant scale $C>0$, \ie, $\phi(\mu(I)) = C\mu(I)$:
{\small{
\begin{equation}
\sum\nolimits_{\phi(I_1)}^{\phi(I_2)} \phi(\mu(I)) = C\sum\nolimits_{I_1}^{I_2} \mu(I).
\label{eq:method_eq4}
\end{equation}
}}
\!\!By combining Eq.~\eqref{eq:intro_eq1},~\eqref{eq:intro_eq2},~\eqref{eq:method_eq3}, and~\eqref{eq:method_eq4}, we have the following ratio equality as our \textsc{EqSim} constraint:
{\small{
\begin{equation}
\frac{s_{11}-s_{12}}{s_{11}-s_{21}}=\frac{s_{22}-s_{21}}{s_{22}-s_{12}}=C = 1.
\label{eq:method_eq5}
\end{equation}
}}
\!Note that $C =1$ can be derived by using the fact that $s_{11}>s_{12}$ and $s_{22}>s_{21}$. By simplifying Eq.~\eqref{eq:method_eq5} further, we have the following two regularizations:
{\small{
\begin{equation}
\begin{split}
&\textsc{EqSim}_\textrm{v1}: s_{12}=s_{21}\\
&\textsc{EqSim}_\textrm{v2}: s_{11}\!-\!s_{12} = s_{22}\!-\!s_{21},~~s_{11}\!-\!s_{21} = s_{22}\!-\!s_{12}.
\end{split}
\label{eq:method_eq6}
\end{equation}
}}
\!\!Note that the viable space of \algnamevv is a subset of \algnamev, because \algnamev is exactly equivalent to Eq.~\eqref{eq:method_eq5} while \algnamevv further requires $s_{11}=s_{22}$.
Empirically, we find that \algnamevv is more suitable to the semantically close pairs $(I_1, T_1)$ and $(I_2, T_2)$; and \algnamev to distant pairs. Figure~\ref{fig:eqsim} illustrates such hybrid training loss within a training batch.
Semantically ``close'' and ``distant'' are determined by the similarity score $s$, where we regard samples with top-$k$ $s$ as ``close'' samples.
For dual encoder VLMs with ITC loss, $s$ is the cosine similarity between image and text features. For fusion encoder VLMs with ITM, $s$ is the scoring output from the ITM head.

In our implementation, we adopt Mean Square Error (MSE) loss to regularize the equation of similarities. In addition, motivated by the hinge loss~\cite{boser1992training,vapnik1999nature}, we utilize a margin parameter $\alpha$ to control the strength of regularization. \textsc{EqSim}$_\textrm{v1}$ can be written as $[||s_{12}-s_{21}||_{2}^{2}-\alpha]_{+}$, where $[x]_{+}=\text{max}(x,0)$ and $||\cdot||_{2}$ denotes the L2 norm. \textsc{EqSim}$_\textrm{v2}$ can be implemented similarly.
In practice, given a retrieval fine-tuning objective $\mathcal{L}_{Ret}$, the final loss can be written as: $\mathcal{L} = \mathcal{L}_{Ret}+\beta \mathcal{L}_{\textsc{Eq}}$, where $\mathcal{L}_{\textsc{Eq}}$ is \algnamev  (\algnamevv) for semantically distant (close) samples, $\beta$ is the balancing factor.
Experiments in Section~\ref{sec:ablation} validate that the hybrid training is better than only using \textsc{EqSim}$_\textrm{v1}$.

\section{Diagnosing VLMs with \benchname}\label{sec:dataset}

We argue that standard VL evaluation kits~\cite{plummer2015flickr30k, Lin2014MicrosoftCC} are too coarse to evaluate the equivariance of VLMs similarity. Existing VLMs can easily distinguish most samples in conventional retrieval benchmarks, \eg, images of a group of people against those with cars, given a caption of ``people standing on the street''.
Therefore, we propose \benchname to focus on visual minimal semantic changes to check whether VLMs can faithfully respond, \ie, the equivariance of the similarity measure in VLMs.
Specifically, \benchname contains 
5 sub-datasets, covering diverse image domains, from real-life scenarios to synthetic well-controlled scenes. And it is designed to stress test VLMs with accurate semantic changes in action, location, and attribution (\eg, color, count and size). Figure~\ref{fig:benchmark} presents an overview of \benchname. 

Next, we introduce our design principle for constructing \benchname. Each sample in \benchname consists of a pair of images $(I_1, I_2)$ and a pair of captions $(T_1, T_2)$. A valid \benchname sample must satisfy:  (1) $T_i$  is preferred to be used as the description for $I_i$;
(2) $I_1$ and $I_2$ are visual-minimally different. 
The former one requires that $\{I_1,T_1\}$ and $\{I_2, T_2\}$ should be semantically distinguishable without confusion, while the latter one limits the extent of the distinction -- ``visual-minimal change''. 
Previous work~\cite{thrush2022winoground} defines ``minimal'' semantic change in the caption space as the same words but in a different order. However, due to the continuity and entanglement of image pixels, ``minimal'' semantic change in visual space is hard to determine. In this paper, we roughly define it as changes in the foreground (\eg, attribute, action, location, \etc) while sharing the same scene and background.

In practice, we source image pairs with ``visual-minimal change'' in two ways: (1) from natural videos and (2) from synthetic engines, where we adopt different construction pipelines, as shown at the bottom of Figure~\ref{fig:benchmark}.
For the former one, we directly leverage the \underline{continuity} of scene changes along the temporal dimension in \emph{natural videos}, which can provide massive image pairs with minimal visual changes. Specifically, we leverage the existing video-language datasets~\cite{zhou2018towards, ji2020action, wang2022geb+} to construct \benchname samples.
To more precisely control the varying component in images, we further explore the 
photo-realistic scene generator (Kubric~\cite{greff2021kubric}) and the open-source diffusion model (Stable Diffusion~\cite{rombach2022high, hertz2022prompt}) to synthetically generate pairs of images by providing two captions that are minimally different from each other.
In what follows, we introduce the construction pipeline for each sub-dataset in detail.

\subsection{Construction from Natural Video}
Let's define a video with caption annotations as $\mathcal{V}=\{I_i, T_i\}_{i=1}^{N}$, where $N$ is the number of sampled frames. 
We assume the ``visual-minimal change'' is naturally guaranteed between any two frames $I_i$ and $I_j$ ideally, where $I_i, I_j \in \mathcal{V}$, $i\neq j$, as we limit the source video to be either short segment~\cite{wang2022geb+,ji2020action} or capturing a fixed scene~\cite{zhou2018towards}.
However, we find that it is hard to ensure the validity of all the video frame pairs in practice. Therefore, we utilize a frame filter to filter out invalid samples automatically for different video sources.
Below, we briefly introduce the dataset construction process and delay the  details to Appendix. 

We construct three sub-datasets based on real images from natural videos, including \textsc{Eq-AG}, \textsc{Eq-GEBC} and \textsc{Eq-YouCook2}. We construct \textbf{\textsc{Eq-AG}} by leveraging the scene graph annotations from Action Genome (AG)~\cite{ji2020action}, which capture detailed changes between objects and their pairwise relationships while action occurs.  
We first use a slot-filling template to translate scene graphs to captions. 
As videos usually come with redundant frames, we avoid the nearly duplicated frames by sampling frames $I_i$ and $I_j$ if and only if at least 2 of 3 pairwise relationships are different.
Furthermore, if $I_j$ is chosen for previous samples, we empirically skip the subsequent 2 frames to $I_{j+2}$.
\textbf{\textsc{Eq-GEBC}} is built on GEBC~\cite{wang2022geb+} that contains captions describing the event before and after an event boundary. We adopt the frames before and after the boundary as our visual minimally different images. Similarly, we avoid temporal redundancy via sparse sampling across multiple boundaries. We construct \textbf{\textsc{Eq-YouCook2}} based on YouCook2~\cite{zhou2018towards}, which is sparsely annotated with captions for each cooking step. We construct the dataset by sampling the middle frame of a short video segment as $I_i$ with its annotated caption as $T_i$. We then apply off-the-shelf object detectors to filter scene changes. Please note that \benchname can be easily extended to other video-language datasets by applying the same construction pipeline.

\subsection{Construction from Synthetic Engine}
Synthetic engine may provide more precise and controllable visual changes in the generated images, to allow more accurate diagnosis in terms of model failure when evaluating with \benchname. We assume that a synthetic engine can faithfully generate images based on a text prompt describing the image content.
Based on this assumption, given a pair of semantic-minimally different captions $T_i$ and $T_j$, we expect the generated $I_i$ and $I_j$ to be correspondingly visual-minimally different. In the following, we briefly introduce the utilized engine and how to construct semantic-minimally different captions for each sub-dataset and leave more details to Appendix.

\textbf{\textsc{Eq-Kubric}} takes advantages of Kubric~\cite{greff2021kubric}, an open-source graphics engine 
to generate photo-realistic scenes. Here we adopt Google Scanned Objects (GSO) for scene construction and categorize caption change into three aspects: \textit{attribute}, \textit{counting}, and \textit{location}. For each aspect, we construct $2000$ image-text pairs by intervening corresponding phrases of sentences while leaving other words unchanged.
\textbf{\textsc{Eq-SD}} is inspired by the recent advances in diffusion models for text-to-image generation~\cite{ramesh2022hierarchical, saharia2022photorealistic, yu2022scaling}. 
We utilize the open-source checkpoint v1.4 of Stable Diffusion (SD) with prompt-to-prompt image editing framework~\cite{hertz2022prompt} to translate two semantic-minimally different captions to a pair of images. 
Specifically, we elaborately design a set of textual semantic-minimal editing: 1) object change (\eg, ``dog''$\to$``cat''); 2) scene change (\eg, + ``in the winter''); 3) attribute change (\eg, + ``with a sunglasses''). Finally, we perform a human evaluation to filter out poor-quality generations.
Notably, we can adopt more rendered objects (\eg, rendered animals) and various synthetic engine (\eg, better generative models) to further extend our \benchname following the proposed pipeline.

\begin{table}[t]
\tablestyle{3pt}{1.2} 
\centering
\resizebox{\linewidth}{!}{
\begin{tabular}{lccccc}
\shline
Dataset &\makecell[c]{\# Testing\\ Samples}  & \makecell[c]{Visual Semantic\\ Change} & \makecell[c]{Pairwise} & \makecell[c]{Domain\\ Diversity}  &Scalability \\ \midrule
Flickr30K~\cite{plummer2015flickr30k} & 1K  & \XSolidBrush  & \XSolidBrush & \XSolidBrush  & \Checkmark\\
COCO~\cite{chen2015microsoft} & 1/5K   & \XSolidBrush & \XSolidBrush & \XSolidBrush & \Checkmark\\
VALSE~\cite{parcalabescu2021valse} & 6795  & \XSolidBrush & \Checkmark & \Checkmark  & \Checkmark\\
Winoground~\cite{thrush2022winoground} & 400  & \XSolidBrush  & \Checkmark & \XSolidBrush  & \XSolidBrush\\ \hline
\textbf{\benchname (Ours)} &250K  & \Checkmark & \Checkmark  & \Checkmark & \Checkmark \\ 
\shline
\end{tabular}}
\caption{Comparison between \benchname and related benchmarks. 
}
\label{tab:intro_bench}
\end{table}

\begin{table*}[t]
\centering
\tablestyle{8.5pt}{1.1} 
\begin{tabular}{lccc|cc|cccccc}
\shline
\multicolumn{1}{l}{\multirow{2}{*}{Method}} & \multicolumn{3}{c|}{Winoground} & \multicolumn{2}{c|}{VALSE} & \multicolumn{3}{c}{F30K Text-to-image Ret.} & \multicolumn{3}{c}{F30K Image-to-text Ret.} \\
\cmidrule(lr){2-4} \cmidrule(lr){5-6}\cmidrule(lr){7-12}
& Text & Image & Group &min($p_c$,$p_f$)  &acc   & R@1 & R@5 & R@10 & R@1 & R@5 & R@10 \\ 
\hline
METER~\cite{dou2021empirical} & 39.25 & 15.75 & 11.99 &25.43  & 54.01  & 79.60 & 94.96 & 97.28 & 90.90 & 98.30 & 99.50 \\
+ FT (F30K)~\cite{dou2021empirical} & 43.50 & 20.75 & 14.75 & 22.41  & 53.06  & 82.22 & 96.34 & 98.36 & 94.30 & \textbf{99.60} & \textbf{99.90} \\
+ \algname &\cellcolor{mygray}\textbf{44.99} &\cellcolor{mygray}\textbf{22.75} &\cellcolor{mygray}\textbf{18.75} &\cellcolor{mygray}\textbf{30.08}  &\cellcolor{mygray}\textbf{53.38} &\cellcolor{mygray}82.16 &\cellcolor{mygray}94.70 &\cellcolor{mygray}96.64 &\cellcolor{mygray}\textbf{95.30} &\cellcolor{mygray}\textbf{99.60} &\cellcolor{mygray}\textbf{99.90} \\ 
\hline
FIBER~\cite{dou2022coarse} & 46.25 & 25.75 & 22.24 & 11.42  & 51.76 & 79.26  &95.70  &97.92  &91.60  &99.50  &99.80  \\
+ FT (F30K)~\cite{dou2022coarse} & 51.24 &  26.49 & 23.00  &20.78  & 54.90 & 81.44 & 96.72 & \textbf{98.48} & 92.90 & 99.50 & \textbf{99.90} \\
+ \algname &\cellcolor{mygray}\textbf{51.49} &\cellcolor{mygray}\textbf{31.49} &\cellcolor{mygray}\textbf{27.50}  &\cellcolor{mygray}\textbf{52.42}  &\cellcolor{mygray}\textbf{58.06}  &\cellcolor{mygray}\textbf{83.56} &\cellcolor{mygray}\textbf{96.78} &\cellcolor{mygray}98.28 &\cellcolor{mygray}\textbf{96.00} &\cellcolor{mygray}\textbf{99.60} &\cellcolor{mygray}\textbf{99.90}   \\   
\shline
\end{tabular}
\caption{Results of \algname on the challenging Winoground~\cite{thrush2022winoground}, VALSE~\cite{parcalabescu2021valse} benchmark and Flickr-30K (F30K)~\cite{plummer2015flickr30k} test split for image-text retrieval. FT and Ret. are short for fine-tuning and retrieval.
}
\label{tab:eqsim_eval}
\end{table*}

\subsection{Comparisons with Other Datasets}
In Table~\ref{tab:intro_bench}, we conduct a direct comparison of \benchname against two widely adopted retrieval benchmarks (Flickr30K~\cite{plummer2015flickr30k} and COCO~\cite{chen2015microsoft}) and two recent datasets with textual-minimal change (VALSE~\cite{parcalabescu2021valse} and Winoground~\cite{thrush2022winoground}) from four aspects.
1) On the dataset characteristics, to the best of our knowledge, \benchname is the first diagnosing benchmark to examine the equivariance of VLMs in terms of \textbf{minimal visual semantic change}.
2) For evaluation setting, \textbf{pairwise} setting asks VLMs to select the correct counterpart within a pair of slightly different samples rather than thousands of very different samples in conventional retrieval datasets. The minimal semantic drift between the pair of samples makes the evaluation of equivariant similarity measure more effective. 
3) For \textbf{domain diversity}, 
our \benchname contains rich visual contents collected from different video domains as well as synthetic domains, as opposed to the common image-text datasets which existing diagnosing kits are built upon.
4) In terms of \textbf{scalability}, 
\benchname is highly scalable as our automatic pipeline can be easily applied to other video-language datasets and synthetic engines, as opposed to manual annotation for building traditional retrieval datasets. While VALSE only focuses on linguistic editing of the captions, \benchname can be further scale up with more diverse visual contents.

\section{Experiments}
We first introduce our experimental setting in Section~\ref{sec:exp-setup}, 
followed by evaluation of \algname on existing benchmarks in Section~\ref{sec:exp_eqsim}.
Section~\ref{sec:exp_eqbench} benchmarks SOTA VLMs on \benchname to show their insensitivity to minimal visual semantic changes, and we further validate \algname on \benchname. Section~\ref{sec:ablation} presents additional ablation studies to examine the design of \algname. 

\subsection{Experimental Setting}
\label{sec:exp-setup}
\noindent\textbf{Training Details}. Recent efforts on diagnosing benchmarks~\cite{thrush2022winoground,parcalabescu2021valse} only provide testing data and directly evaluate models after VL pre-training. The low performance reported on these benchmarks can mainly be attributed to two factors: 1) the inherent weaknesses of VLMs, \eg, non-equivariant similarity measure; and 2) the domain gap between training and testing. To better validate the effectiveness of our method, we fine-tune the VLMs on limited image-text pairs from conventional retrieval dataset Flickr30K~\cite{plummer2015flickr30k} with or without the regularization term of \algname, and then test the fine-tuned VLMs on the challenging Winoground~\cite{thrush2022winoground}, VALSE~\cite{parcalabescu2021valse} and our \benchname. Under a fair comparison, we argue that the absolute performance improvements from \algname thus would suggest that the gain is entirely from the remedy of model weaknesses.

To validate the effectiveness and genraliazability of our proposed method, we apply \algname to two SOTA end-to-end methods with different architectures and retrieval losses. Specifically, FIBER~\cite{dou2022coarse} supports the dual encoder with ITC loss for fast retrieval, which computes similarities for $N^2$ image-text pairs with only $O(N)$ forwarding. In contrast, the SOTA fusion-encoder model METER~\cite{dou2021empirical}, optimized with ITM task during pre-training, computes the similarity by forwarding the concatenation of each pair of image and text, resulting in $O(N^2)$ time complexity. Fine-tuning details for each model can be found in Appendix.

\noindent\textbf{Evaluation Metric}. On \textbf{Winoground}~\cite{thrush2022winoground}, given two image-text pairs $\{I_1,T_1\}$ and $\{I_2,T_2\}$, a VLM measures similarity $s_{ij}$  between image $I_i$ and text $T_j$ $(i,j\in\{0,1\}, i\neq j)$. Three metrics are computed based on $s_{ij}$: 1) \emph{Text score} measures whether the model can select the correct text for a given image. The model wins one point if $s_{ii} > s_{ij}$. 2) \emph{Image score} 
 evaluates if VLMs can select the correct image for a given text and the model wins one point when $s_{ii} > s_{ji}$. 3) \emph{Group score} combines the previous two, such that the VLMs win one point if and only if both text score and image score are 1, meaning the following condition must be satisfied: $s_{ii} > s_{ij}$ and $s_{ii} > s_{ji}$.
On \textbf{VALSE}~\cite{parcalabescu2021valse}, the two image-text pairs share a common image, \ie, $\{I_1, T_1\}$ (correct) and $\{I_1, T_2\}$ (foil). We follow~\cite{parcalabescu2021valse} to report the following metrics: 1) \emph{acc} is the overall accuracy on both correct and foil image-text pairs; and 2) \emph{min($p_c,p_f$)} is the minimum of precision $p_c$ and foil precision $p_f$, where $p_c$ ($p_f$) measures how well models identify the correct (foil) pair.
We also report performance on the conventional image-text retrieval task, where recall R@K (K=1,5,10) is used as the evaluation metric.

\begin{table*}[t]
\centering

\tablestyle{3pt}{1.02} 
\begin{tabular}{l|ccccccccc|cccccc|l}
\shline
\multirow{3}{*}{Method} & \multicolumn{9}{c|}{Natural Subsets}  & \multicolumn{6}{c|}{Synthetic Subsets}& \multirow{3}{*}{Avg}\\
\cmidrule(lr){2-10} \cmidrule(lr){11-16}
 & \multicolumn{3}{c}{\textsc{Eq-YouCook2}} & \multicolumn{3}{c}{\textsc{Eq-GEBC}} & \multicolumn{3}{c|}{\textsc{Eq-AG}} & \multicolumn{3}{c}{\textsc{Eq-Kubric}} & \multicolumn{3}{c|}{\textsc{Eq-SD}} &  \\
\cmidrule(lr){2-4} \cmidrule(lr){5-7} \cmidrule(lr){8-10} \cmidrule(lr){11-13} \cmidrule(lr){14-16}
   & Text & Image & Group & Text & Image & Group & Text & Image & Group & Text & Image & Group & Text & Image & Group &  \\ 
\hline
LXMERT~\cite{tan-bansal-2019-lxmert} &13.96  &11.98  &4.55  &13.56  &12.73  &4.19  &18.17  &9.02  &4.46  &18.50  &15.35  &7.26  &11.16  &6.15  &1.98   &10.20 \\
ViLBERT~\cite{lu2019vilbert} &14.78  &12.75  &5.18  &14.67  &12.64  &4.82  &17.43  &8.36  &3.89  &17.55 &18.44 &8.13  &12.37  &7.37  & 2.78  &10.74 \\
\hline
CLIP (RN-50)~\cite{radford2021learning} & 47.72 & 47.99 & 34.05 & 10.80 & 18.03 & 3.97 & 14.52 & 10.44 & 3.50 & 21.33 & 21.93 & 9.75  &90.09 &  85.92 & 79.11 &  33.28\\
CLIP (ViT-B/32)~\cite{radford2021learning} & 49.48 & 51.10 & 36.50 & 12.57 & 20.12 & 4.47 & 13.91 & 8.72 & 3.32 & 20.56 & 21.29 & 9.66 & 89.16 & 86.05 &78.98 &  33.73\\
FLAVA~\cite{singh2022flava} & 51.66 & 54.78 & 39.68 &12.24  &16.81  &5.07  & 6.59 & 13.47 & 2.15 & 28.88 & 28.18 & 15.90 &  79.64 & 84.47 & 71.10 &  34.04\\
ViLT~\cite{kim2021vilt} & 44.61 & 46.69 & 31.74 & 14.72 & 16.70 & 5.62 & 15.37 & 9.89 & 3.45 & 31.23 & 27.00 & 17.90 & 80.37 & 79.04 & 68.93 &  32.88\\
ALBEF~\cite{li2021align} & 57.01 & 58.04 & 44.90 & 13.56 & 19.63 & 5.89 & 11.28 & 15.17 & 3.93 & 29.87 & 30.18  & 18.58 & 88.96 & 90.41 & 83.07 &  38.03\\
BLIP~\cite{li2022blip} & 59.22 & 58.36 & 46.31 &  15.87 & 19.79 & 7.27 & 19.76 & 13.87 & 6.31 & 29.38 & 32.25 &  18.73 & 85.39 & 85.52 & 77.13  &  38.34\\ 
\hline
METER~\cite{dou2021empirical} & 52.18 & 49.42 & 36.81 & 20.95 & 18.19 & 6.95 & 28.70 & 15.80 & 7.88  & \textbf{44.28} & 35.20 & 27.26  & \textbf{89.62} & \textbf{84.93} & \textbf{79.44}  & 39.84\\
 + FT (F30K)~\cite{dou2021empirical} & 52.68 & 48.31 & 36.52 & 18.08 & 19.85 & 7.33 & \textbf{29.50} & 16.30 & 8.12 & 41.11 & 34.59 & 24.33  & 86.64 & 84.46 & 77.46  &  39.02\\
 + \algname &\cellcolor{mygray}\textbf{54.12}  &\cellcolor{mygray}\textbf{53.12}  &\cellcolor{mygray}\textbf{40.29} &\cellcolor{mygray}\textbf{24.20}  &\cellcolor{mygray}\textbf{26.02}  &\cellcolor{mygray}\textbf{11.69} &\cellcolor{mygray}28.85  &\cellcolor{mygray}\textbf{20.09}  &\cellcolor{mygray}\textbf{10.76} &\cellcolor{mygray}\underline{43.68}  &\cellcolor{mygray}\textbf{39.08}  &\cellcolor{mygray}\textbf{28.42}  &\cellcolor{mygray}\underline{88.04}&\cellcolor{mygray}84.07 &\cellcolor{mygray}\underline{77.79}  &\cellcolor{mygray}\textbf{42.28}  \\ 
\hline
FIBER~\cite{dou2022coarse} & 52.04 & 50.84 & 38.32 & \textbf{25.19} & 22.66 & \textbf{11.08} & \textbf{32.49} & \textbf{24.05} & \textbf{13.70} & 47.94 & 45.60 & 33.53 & 86.05 & \textbf{88.63} & 79.97  &  44.86\\
+ FT (F30K)~\cite{dou2022coarse} & 57.70 & 56.46 & 44.33 & 18.24 & 21.33 & 8.54 & 26.99 & 18.69 & 9.24 & 50.31 & 46.06 & 34.66 & 90.48 & 86.64 & \textbf{81.29}  &   43.40\\
 + \algname &\cellcolor{mygray}\textbf{58.26}  &\cellcolor{mygray}\textbf{57.10}  &\cellcolor{mygray}\textbf{45.10} &\cellcolor{mygray}\underline{21.55} &\cellcolor{mygray}\textbf{26.07} &\cellcolor{mygray}\underline{10.58} &\cellcolor{mygray}\underline{29.93}  &\cellcolor{mygray}\underline{23.42}  &\cellcolor{mygray}\underline{12.64}  &\cellcolor{mygray}\textbf{51.90}  & \cellcolor{mygray}\textbf{48.40} &\cellcolor{mygray}\textbf{37.38}  &\cellcolor{mygray}\textbf{90.81} & \cellcolor{mygray}85.98 &\cellcolor{mygray}80.70  &\cellcolor{mygray}\textbf{45.32} \\ 
\shline
\end{tabular}
\caption{Results on \benchname. 
Rows highlighted in gray are results with our \algname. Numbers in bold, underline respectively represent the best results and the inferior results compared to pre-training but better than fine-tuning baseline. Avg denotes the average over all scores.
}
\label{tab:eqbench}
\end{table*}
\subsection{Evaluation of \algname}\label{sec:exp_eqsim}

In Table~\ref{tab:eqsim_eval}, we compare model performance under three settings: ($i$) direct evaluation after pre-training (the first rows of each block); ($ii$) standard fine-tuning (FT) on Flickr30K training data (the second rows of each block); and ($iii$) fine-tuning with \algname regularization (the third rows of each block). 
We observe that standard fine-tuning can somewhat bring a little performance improvement on both Winoground and VALSE benchmarks, indicating that some domain overlap between Flickr30K training data and testing samples.
It is difficult to entirely rule out the domain influence, but comparing fine-tuning with \algname against standard fine-tuning, 
our method brings consistent and significant performance improvements on both of the challenging Winoground and VALSE across METER and FIBER models. Specifically, \algname improves the group score over standard fine-tuning by $4\%$ for METER and $4.5\%$ for FIBER on Winoground, respectively. 
While for VALSE, the performance improvement on min$(p_c,p_f)$ is as large as $31.6\%$, further validating the effectiveness of our \algname. 
In addition, we observe that the equivariance regularization from \algname does not sacrifice retrieval performance. On Flickr30K, \algname can mostly retain the retrieval performance, and sometimes even yield performance gain, \eg, $3.1\%$ on R@1 for image-to-text retrieval with FIBER.

\begin{table}[t]
\centering
\tablestyle{6pt}{0.98} 
\begin{tabular}{lccc}
\shline
\multirow{2}{*}{Method} & \multicolumn{3}{c}{\textsc{Eq-Kubric}}\\
\cmidrule(lr){2-4}
 & Location & Counting & Attribute  \\
 \hline
LXMERT &2.15  &1.89  &17.90    \\
ViLBERT~\cite{lu2019vilbert} &1.98  &2.03  &20.40     \\
\hline
CLIP (RN-50)~\cite{radford2021learning} &  1.25 &  5.20 & 22.80  \\
CLIP (ViT-B/32)~\cite{radford2021learning} &   0.75  & 5.80 & 22.44 \\
FLAVA~\cite{singh2022flava}  &  1.00  & 7.35  & 39.35   \\
ViLT~\cite{kim2021vilt} & 1.95 &  6.19 &45.55   \\
ALBEF~\cite{li2021align} & 1.25 &  9.49 &  44.99   \\
BLIP~\cite{li2022blip} &  1.15 &  10.60 & 44.44   \\ 
\hline
METER~\cite{dou2021empirical} & 3.59 &  15.29 &  \textbf{62.90}   \\
+ FT (F30K)~\cite{dou2021empirical} &  2.40 &  17.49  & 53.10   \\
+ \algname  &\cellcolor{mygray}\textbf{3.80}  &\cellcolor{mygray}\textbf{23.85} &\cellcolor{mygray}\underline{57.60} \\ 
\hline
FIBER~\cite{dou2022coarse} &  \textbf{11.34} &  19.65 &  69.59 \\
+ FT (F30K)~\cite{dou2022coarse} &  8.95 &  28.49 & 66.54 \\
+ \algname &\cellcolor{mygray}\underline{11.05} &\cellcolor{mygray}\textbf{30.90} &\cellcolor{mygray}\textbf{70.20} \\ 
\shline
\end{tabular}
\caption{Detailed results on \textsc{Eq-Kubric}. We report group score on three splits of \textsc{Eq-Kubric}, capturing the visual semantic changes in Location, Counting and Attribute. 
} 
\label{tab:kubric}
\end{table}

\subsection{Benchmarking VLMs with \benchname}\label{sec:exp_eqbench}
We evaluate a wide range of VLMs with different configurations on \benchname in a zero-shot manner, to examine the equivariance of their similarity measures for distinguishing visually-minimal different samples. We consider representative VLMs, including ($i$) LXMERT~\cite{tan-bansal-2019-lxmert}, ViLBERT~\cite{lu2019vilbert} for OD-Based models; and ($ii$) CLIP~\cite{radford2021learning} variants,  FLAVA~\cite{singh2021flava}, ViLT~\cite{kim2021vilt}, ALBEF~\cite{li2021align}, BLIP~\cite{li2022blip} and METER~\cite{dou2021empirical} and FIBER~\cite{dou2022coarse}  as prominent examples of end-to-end SOTA methods.
Full results on more VLMs can be found in Appendix~\ref{supp_sec:eqben_results}.  For evaluation metrics, we adopt text score, image score and group score to compare model performance, similar to Winoground~\cite{thrush2022winoground}.

Table~\ref{tab:eqbench} presents the evaluation results of existing VLMs on \benchname and we summarize our observations below.
\vspace{-3pt}
\begin{itemize}[leftmargin=*]
\setlength\itemsep{-2pt}
    \item Regardless of the subsets, end-to-end VLMs generally achieve better performance as it is not constrained by the fixed visual representation from a pre-trained object detector~\cite{ren2015faster}, as in OD-based methods.
    \item Among all subsets, VLMs obtain evidently higher performance on \textsc{Eq-SD}. 
    The stable diffusion model~\cite{rombach2022high} is pre-trained on  similar VL corpus to these VLMs. Hence, the generated images can be biased towards the same underlying data distribution, much easier for VLMs to tell the differences.
    Besides, the generated images maybe visually minimally different to human eyes, but it is unclear whether in the pixel space, they are minimally different \wrt the model input. 
    It is worth noting that LXMERT and ViLBERT are the exception due to the totally different distribution with the off-the-shelf object detector.
    \item Interestingly, a larger pre-training corpus (\eg, CLIP~\cite{radford2021learning} and FLAVA~\cite{singh2021flava}) does not always guarantee better results. This 
    implies training loss may be more critical in learning equivariant similarity measure.

    \item In Table~\ref{tab:kubric}, we further conduct a fine-grained examination with the synthetic subset \textsc{EQ-Kuric}, where we focus on specific visual changes in location, counting and attribute.
    VLMs fail substantially in terms of location and counting, while being sensitive to attribute changes. Similar findings are also observed by ~\cite{thrush2022winoground, parcalabescu2021valse} from the text side. 
\end{itemize}
\vspace{-3pt}
We again equip the two strong baseline models (METER and FIBER) with \algname and fine-tune on Flickr30K. As \benchname covers diverse domains, standard fine-tuning on Flickr30K can hardly improve or even hurt model performance, compared with direct evaluation after pre-training (with $-0.62\%$ and $-1.46\%$ performance drop for METER and FIBER, respectively). However, by enforcing equivariant constraint with \algname, we observe significant performance improvements than standard fine-tuning, with an absolute gain of $3.26\%$ for METER and $1.92\%$ for FIBER.

\begin{table}[t]
\centering

\tablestyle{3.7pt}{1.15} 
\resizebox{\linewidth}{!}{
\begin{tabular}{llccccc}
\shline
FT Data & Method & \textsc{Eq-AG} & \textsc{Eq-Y.} & \textsc{Eq-G.} & Wino. & Avg \\ \hline
\multirow{2}{*}{F30K~\cite{plummer2015flickr30k}} & FT & 9.24 & 44.33 & 8.54 & 23.00 & 21.28 \\
 & + \algname & \cellcolor{mygray}\textbf{12.64} & \cellcolor{mygray}\textbf{45.10} & \cellcolor{mygray}\textbf{10.58} & \cellcolor{mygray}\textbf{27.50} & \cellcolor{mygray}\textbf{23.96} \\ 
 \hline
\multirow{2}{*}{COCO~\cite{chen2015microsoft}} & FT & 10.14 & 42.90 & 8.93 & 21.50 & 20.87 \\
 & + \algname & \cellcolor{mygray}\textbf{12.52} & \cellcolor{mygray}\textbf{45.68} & \cellcolor{mygray}\textbf{9.37} & \cellcolor{mygray}\textbf{25.75} & \cellcolor{mygray}\textbf{23.33} \\ \hline
\multirow{2}{*}{F30K + COCO} & FT & 9.96 & 43.81 & 8.93 & 22.75 & 21.36 \\
 & + \algname & \cellcolor{mygray}\textbf{11.98} & \cellcolor{mygray}\textbf{45.80} & \cellcolor{mygray}\textbf{10.47} & \cellcolor{mygray}\textbf{26.50} & \cellcolor{mygray}\textbf{23.69} \\ 
  \hline
\multirow{2}{*}{4M$^{\dagger}$} & FT & 10.49 & 40.95 & 7.49 & 20.99 & 19.98 \\
 & + \algname & \cellcolor{mygray}\textbf{12.78} & \cellcolor{mygray}\textbf{40.96} & \cellcolor{mygray}\textbf{9.81} & \cellcolor{mygray}\textbf{21.25} & \cellcolor{mygray}\textbf{21.20} \\
\shline
\end{tabular}
}
\caption{Group accuracy (\%) of fine-tuning (FT) and \algname based on FIBER on different FT data corpus. 4M data denotes the commonly used pre-training data with about 4M images, including COCO, Visual Genome~\cite{krishna2016visual}, Conceptual Captions~\cite{sharma2018conceptual} and SBU~\cite{ordonez2011im2text}. Y., G., Wino. are the short for YouCook2, GEBC and Winoground. $\dagger$ The model is fine-tuned for 10K steps.
}
\label{tab:datasize}
\end{table}

\subsection{Ablation Study}
\label{sec:ablation}

In this section, we conduct ablation studies to validate the scalability, design and effectiveness of \algname in terms of enforcing equivariant similarity. 

\noindent\textbf{Scalability of \algname}. Table~\ref{tab:datasize} evaluates the scalability of \algname and standard fine-tuning baseline on the natural subsets of \benchname and Winoground by gradually including more training data. Under the same fine-tuning data, \algname achieves consistent and significant improvements ($2\%$ - $3\%$) over the baseline. Interestingly, there is no remarkable correlation between the corpus size and model performance. This may be due to the distribution of standard VL data is far away from that of \benchname and Winoground. 
Note that for the 4M experiment, we fine-tune the models for 10K steps due to computational constraints. Our results demonstrate the potential of \algname to benefit VL pre-training on large-scale data.
Additionally, we validate the generalizability of \algname in other relevant downstream tasks. Further details are provided in Appendix~\ref{supp_sec:generalization}.

\noindent\textbf{Ablation on \algname design}.
Table~\ref{tab:abla_arch} compares \algname against the four ablated instances on \textsc{Eq-Kubric} and Winoground~\cite{thrush2022winoground}, including 1) fine-tuning with hard negative sampling (HardNeg);  
2) applying \algnamev to all samples in the training batch (\algnamev-all); 3) applying \algnamevv to all samples in the training batch (\algnamevv-all); and 4) applying \algnamevv for only semantically close samples (\algnamevv-close). The final \algname is equivalent to \algnamev-all + \algnamevv-close,
which achieves the best performance. Notably, enforcing \algnamevv on all (\algnamevv-all) even degrades the performance by -0.58\% on average, compared to applying only to semantically close samples (\algnamevv-close). This validates our claim in Section~\ref{sec:method} that \algnamevv is better suited for semantically close samples.

\begin{figure}[t]
    \centering
\hspace{-15pt}
    \includegraphics[width=.5\textwidth]{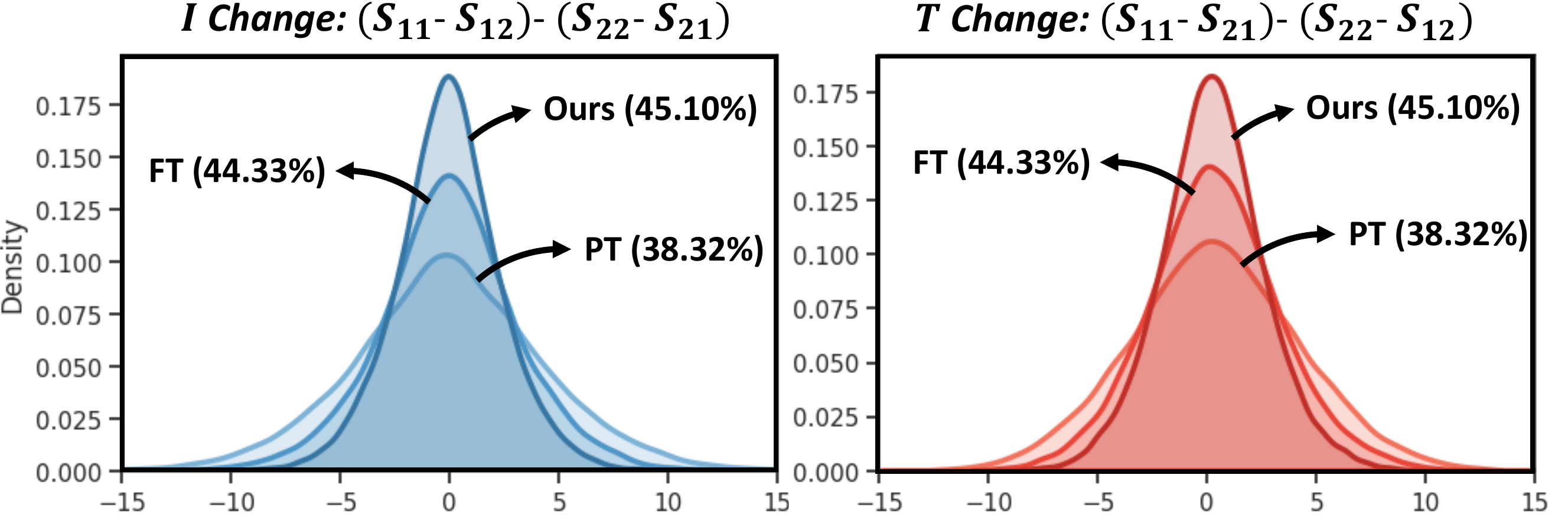}
    \caption{The equivariant score of three FIBER~\cite{dou2022coarse} variant models: Pre-trained (PT), Fine-tuned (FT) and Ours (\algname) on \textsc{Eq-YouCook2}. The equivariant score is defined by the derivation of \algnamevv. The tighter distribution curve, the better equivariant similarity measure.}
    \label{fig:intro_hist}
\end{figure}

\noindent\textbf{Validation of equivariance via \algname}.
Given the similarity scores $s$ calculated by a VLM, we can define the equivariance score as the derivation of \algnamevv (headline of Figure~\ref{fig:intro_hist}) to measure the degree of equivariance (the smaller, the better). In Figure~\ref{fig:intro_hist}, we plot the distribution of \algnamevv values across all samples in \textsc{EQ-Youcook2} dataset for FIBER~\cite{dou2022coarse} and its variants, attached with their group scores. A tighter curve indicates smaller derivation, hence better equivariance similarity measure. 
Full results on other \benchname subset are presented in Appendix~\ref{supp_sec:distribution_curve}.
Compared with pre-training only (PT), fine-tuning on Flickr30K (FT) can improve the group score  while being more equivariant in the similarity measure.
Adding \algname (Ours) obtains additional improvements on both similarity equivariance and group score, indicating \algname indeed enforces equivariant similarity measure. 
Additionally, due to the space limitation, we leave more visualizations in Appendix~\ref{supp_sec:visualization}.

\begin{table}[t]
\centering
\tablestyle{4pt}{1.1} 
\begin{tabular}{lccccc}
\shline
 \multirow{2}{*}{Method} &\multicolumn{3}{c}{\textsc{Eq-Kubric}}   &\multirow{2}{*}{Wino.}  & \multirow{2}{*}{Avg} \\ \cmidrule(lr){2-4}
& Location & Counting & Attribute & \\ \midrule
 FT (F30K) & 8.95 & 28.49 & 66.54 & 22.24  &31.55 \\
 + HardNeg & 10.89 & 29.49 & 67.69 & 27.00  & 33.77 \\ 
 + \algnamev-all & 9.79 & 29.94 & 68.75 & 26.49  & 33.74 \\ 
 + \algnamevv-all & 10.25 & 29.05 & 68.30 & 25.49  & 33.27 \\ 
 + \algnamevv-close & \textbf{11.15} & 29.25 & 69.25 & 25.75  & 33.85 \\  \hline
 + \algname & 11.05 & \textbf{30.90} & \textbf{70.20} & \textbf{27.50}  &\textbf{34.91} \\ 
\shline
\end{tabular}
\caption{Ablation studies of the loss design for our \algname on \textsc{Eq-Kubric}  and Winoground (Wino.) using group score (\%).}
\label{tab:abla_arch}
\end{table}

\subsection{Pilot Study of MLLM on \benchname}
\label{sec:mllm}
Powered by the remarkable capabilities of the large language model (LLM), the community has witnessed an emergent interest in developing Multimodal Large Language Model (MLLM)~\cite{zhu2023minigpt, liu2023visual, gong2023multimodal} very recently. Instead of accepting the pure text as the input, MLLM additionally sees the image and provides the response, which can be regarded as another line of VLMs.
Here we conduct a pilot study of the performance of MLLM on our \benchname. 
We adopt LLaVa-7B~\cite{liu2023visual} as our base model with Vicuna as the LLM backend. Given two matched image-text pairs $\{I_1,T_1\}$ and $\{I_2,T_2\}$, we concatenate $I_1$ and $I_2$ horizontally as the single input image. We build the question prompt with the template: ``\textit{There are two images (left and right). Now you have two captions: caption 1: $\{T_1\}$; caption 2: $\{T_2\}$. Please indicate which caption corresponds to the left image and which caption corresponds to the right one. The answer should follow the format: "\#index for the left image; \#index for the right image". For example, "1;2" represents that caption 1 corresponds to image left.}''
Since it is hard to reformat the MLLM free-form textual output to the label space, we randomly collect 20 samples from each subset of \benchname and manually compare the MLLM output and the ground-truth label. The results are shown in Table~\ref{tab:supp_mllm}.
Interestingly, by comparing two rows, we can find that the performance of MLLM is quite sensitive to the order of the input caption $T_1$ and $T_2$ ($\sim 90\%$ v.s $\sim 0\%$). This indicates that the MLLM does NOT truly understand how to distinguish two semantically similar image-text pairs but just follows the given sequence of the captions.

\begin{table}[t!]
\centering
\tablestyle{3.7pt}{1.15} 
\resizebox{\linewidth}{!}{
\begin{tabular}{cccccc}
\shline
Caption Order & \textsc{Eq-AG} & \textsc{Eq-Y.} & \textsc{Eq-G.} & \textsc{Eq-K.} & \textsc{Eq-SD} \\ \hline
$T_1, T_2$ & 95.00 & 90.00 & 90.00 & 88.33 & 90.00 \\
$T_2, T_1$ & 0.00 & 0.00 & 0.00 & 1.66 & 0.00 \\ 
\shline
\end{tabular}
}
\caption{Group accuracy (\%) of LLaVa on different FT data corpus with different caption input order.  Y., G., K. are the short for YouCook2, GEBC and Kubric.
}
\label{tab:supp_mllm}
\end{table}

\section{Conclusion}
In this study, we investigated the non-equivariant similarity issue in VLMs, hidden behind their excellent performances on standard evaluation benchmarks.
To address this issue, we proposed Equivariance Similarity Learning (\algname), an elegant and effective regularization method that can be easily integrated into the fine-tuning process of existing VLMs. 
Meanwhile, to better diagnose the equivariance of VLMs, we further introduced a new challenging benchmark \benchname, the first to focus on ``visual-minimal change''.
Our proposed \algname is backed by the strong results on both challenging benchmarks (\eg, Winoground, VALSE, \benchname) and the conventional Flickr30K dataset.
In future work, we plan to explore the application of \algname in VL pre-training and instruction tuning.
\noindent\textbf{Acknowledgement.} We thank Ziyi Dou, Xuejiao Zhao for valuable discussions and help, and all 
anonymous reviewers for constructive suggestions. This work is partly supported by AI Singapore AISG2-RP-2021-022.

\clearpage
\vspace{0.2in}
\appendix
\noindent\textbf{\Large Appendix}
\vspace{0.1in}

The Appendix is organized as follows:

\begin{itemize}
    \item Section~\ref{sec:results} includes full illustrations or more experimental results on \benchname and detailed analysis of \algname. 
    \item Section~\ref{sec:data} provides construction details for \benchname.
    \item Section~\ref{sec:implement} presents the implementation details of \algname.
    \item Section~\ref{sec:example} visualizes more examples in \benchname.
\end{itemize}

\section{More Results}
\label{sec:results}

In this section, we include full illustrations and additional experimental results, due to the space limitation of the main paper.

\begin{figure*}[t!]
    \centering
    \includegraphics[width=.95\textwidth]{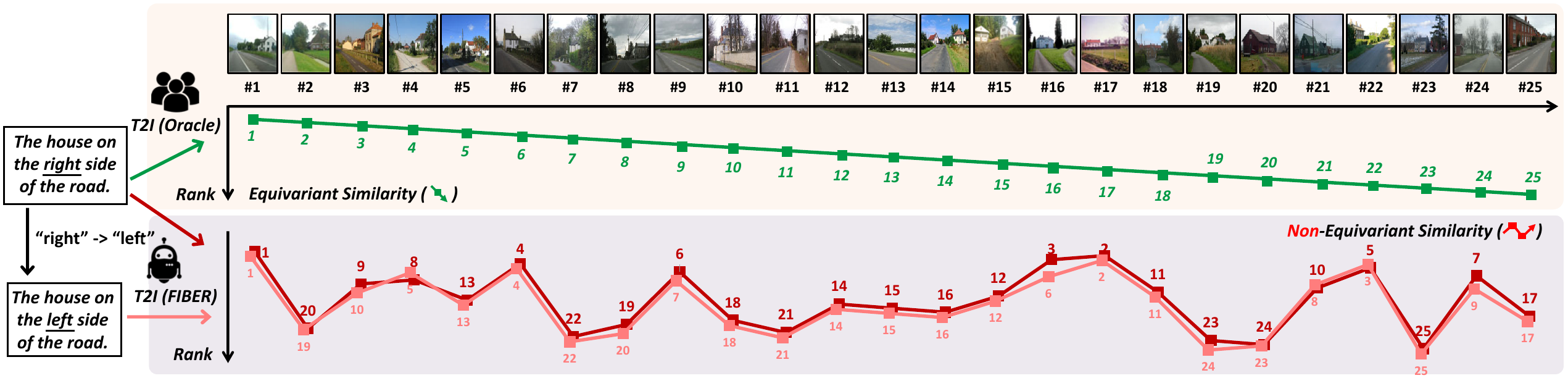}
    \caption{Full ranking results of Figure~\ref{fig:intro_fig}.
    }
    \label{fig:full_rank}
\end{figure*}

\subsection{Full Ranking Results of Figure~\ref{fig:intro_fig}}
\label{supp_sec:full_rank}

In Figure~\ref{fig:intro_fig} of the main paper, we perform a toy experiment on LAION400M to compare the similarity measure of FIBER and the human oracle. Due to the space limitation, we only show partial ranking results in the main paper. Here we illustrate the full ranking in Figure~\ref{fig:full_rank}. 
With the full ranking results, the observation we summarize in the main paper becomes more clear. That is,  the similarity changes in
FIBER do not faithfully reflect the semantic changes in images (\#1 $\rightarrow$ \#25) or text queries ``righ'' $\rightarrow$ ``left”).

\subsection{Retrieval Results on COCO dataset}

We report the retrieval performance of FIBER~\cite{dou2022coarse} variants on COCO~\cite{chen2015microsoft} 5K test split in Table~\ref{tab:supp_indomain}. We observe similar trends on COCO to that on Flickr30K in Table~\ref{tab:eqsim_eval}. The results suggest the effectiveness of the proposed \algname, which brings large performance gain across all metrics. 
\begin{table}[h]
\centering
\tablestyle{4pt}{1.2} 
\begin{tabular}{lcccccc}
\shline
\multicolumn{1}{l}{\multirow{2}{*}{Method}} & \multicolumn{3}{c}{Text-to-Image Ret.} & \multicolumn{3}{c}{Image-to-Text Ret.} \\
\cmidrule(lr){2-4} \cmidrule(lr){5-7}
& R@1 & R@5 & R@10 & R@1 & R@5 & R@10 \\ \hline
FIBER~\cite{dou2022coarse} &55.19  &81.49  &88.89  &73.39  &92.59  &96.41  \\
+ FT (COCO)~\cite{dou2022coarse} & 59.31 & 83.73 & 90.43 & 75.88 & 93.92 & 96.79 \\
+ \algname  & \textbf{62.55} & \textbf{85.36} & \textbf{91.35} & \textbf{80.16} & \textbf{95.44} & \textbf{97.69}      
\\ 
\shline
\end{tabular}
\caption{Image-text retrieval results on COCO~\cite{chen2015microsoft} 5K test set. ``Ret.'' denotes retrieval. Please note that for computation efficiency and fair comparison, we set the image resolution as $288\times288$ during fine-tuning. 
}
\label{tab:supp_indomain}
\end{table}

\subsection{Full Results of Table~\ref{tab:datasize} and Table~\ref{tab:abla_arch}}

We show the full results of ablation studies in Table~\ref{tab:supp_datasize} and Table~\ref{tab:supp_abla_arch}, with group scores across all 5 subsets of \benchname and Winoground.
The observation is similar to the main paper. From Table~\ref{tab:supp_datasize}, we can find that \algname is scalable in terms of training data, showing the potential to benefit VL pre-training. The solitary exception happens on \textsc{Eq-SD}, where \algname cannot consistently obtain the improvements. We hypothesize this is probably because \textsc{Eq-SD} is biased towards the same underlying distribution with the VLMs, as discussed in the main paper. 
With Table~\ref{tab:supp_abla_arch}, we can find that \algname (the hybrid combination of  \algnamev-all and  \algnamevv-close) is the best-performing one, which validates our claim in Section~\ref{sec:method}.
Meanwhile, \algnamev-all and \algnamevv-close also achieve good results (compared with \algnamevv-all), where both of them are supported by the claim in Section~\ref{sec:method}.

\begin{figure*}[t]
    \centering
    \includegraphics[width=.95\textwidth]{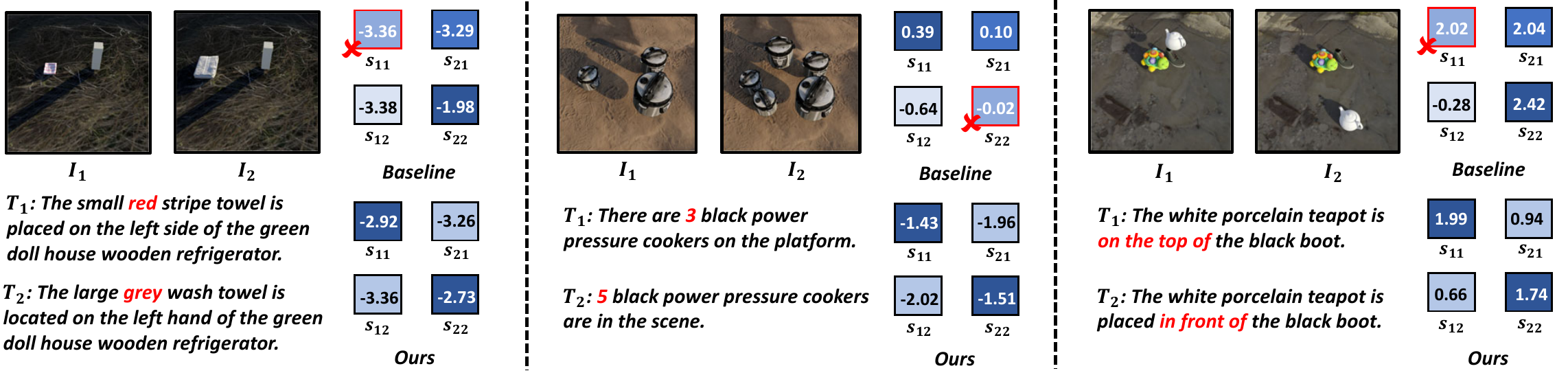}
    \caption{Visualization of similarity scores ($s_{ij}$) on specific examples of \benchname. $s_{ij}$ is the similarity score for ($I_{i},T_{j}$). Darker color indicates larger similarity. The red cross represents the inferior similarity score leading to the wrong matching result.
    }
    \label{fig:sup_vis_eqben}
\end{figure*}

\begin{table*}[t]
\centering

\tablestyle{5pt}{1.1} 

\begin{tabular}{llccccccccc}
\shline
 \multirow{2}{*}{FT Data} &\multirow{2}{*}{Method}  &\multirow{2}{*}{\textsc{Eq-AG}}  &\multirow{2}{*}{\textsc{Eq-YouCook2}}  &\multirow{2}{*}{\textsc{Eq-GEBC}}  &\multicolumn{3}{c}{\textsc{Eq-Kubric}}  &\multirow{2}{*}{\textsc{Eq-SD}} &\multirow{2}{*}{Winoground} & \multirow{2}{*}{Avg} \\ \cmidrule(lr){6-8}
 &   &   &  &   &Location  & Counting  &Attribute &  &  & \\ \hline
\multirow{2}{*}{F30K~\cite{plummer2015flickr30k}} & FT & 9.24 & 44.33 & 8.54  &8.95  &28.49  &66.54  &\textbf{81.29}  & 23.00 &33.79  \\
 & + \algname & \cellcolor{mygray}\textbf{12.64} & \cellcolor{mygray}\textbf{45.10} & \cellcolor{mygray}\textbf{10.58}  &\cellcolor{mygray}\textbf{11.05}  &\cellcolor{mygray}\textbf{30.90}  &\cellcolor{mygray}\textbf{70.20}  &\cellcolor{mygray}80.70  & \cellcolor{mygray}\textbf{27.50} & \cellcolor{mygray}\textbf{36.08} \\ 
 \hline
\multirow{2}{*}{COCO~\cite{chen2015microsoft}} & FT & 10.14 & 42.90 & 8.93  &8.60  &\textbf{26.40}  &66.60  &77.13  & 21.50  &32.77  \\
 & + \algname & \cellcolor{mygray}\textbf{12.52} & \cellcolor{mygray}\textbf{45.68} & \cellcolor{mygray}\textbf{9.37}  &\cellcolor{mygray}\textbf{11.05}  &\cellcolor{mygray}26.15  &\cellcolor{mygray}\textbf{70.75}  &\cellcolor{mygray}\textbf{77.46}   & \cellcolor{mygray}\textbf{25.75} & \cellcolor{mygray}\textbf{34.84} \\ \hline
\multirow{2}{*}{F30K + COCO} & FT & 9.96 & 43.81 & 8.93  &6.94  &28.09  &64.34  & 79.90 & 22.75 &33.09  \\
 & + \algname & \cellcolor{mygray}\textbf{11.98} & \cellcolor{mygray}\textbf{45.80} & \cellcolor{mygray}\textbf{10.47}  &\cellcolor{mygray}\textbf{10.89}  &\cellcolor{mygray}\textbf{31.15}  &\cellcolor{mygray}\textbf{70.20}  &\cellcolor{mygray}\textbf{82.62}  & \cellcolor{mygray}\textbf{26.50} & \cellcolor{mygray}\textbf{36.20} \\ 
  \hline
\multirow{2}{*}{4M$^{\dagger}$} & FT & 10.49 & 40.95 & 7.49  &3.25  &18.09  &65.20  &\textbf{80.63}  & 20.99 &30.88  \\
 & + \algname & \cellcolor{mygray}\textbf{12.78} & \cellcolor{mygray}\textbf{40.96} & \cellcolor{mygray}\textbf{9.81}  &\cellcolor{mygray}\textbf{9.20}  &\cellcolor{mygray}\textbf{21.25}  &\cellcolor{mygray}\textbf{67.90}  &\cellcolor{mygray}79.77  & \cellcolor{mygray}\textbf{21.25} & \cellcolor{mygray}\textbf{32.86} \\
\shline
\end{tabular}
\caption{Full results of Table~\ref{tab:datasize}.
}
\label{tab:supp_datasize}
\end{table*}

\begin{table*}[t]
\centering
\tablestyle{5pt}{1.1} 
\begin{tabular}{lccccccccc}
\shline
 \multirow{2}{*}{Method}  &\multirow{2}{*}{\textsc{Eq-AG}}  &\multirow{2}{*}{\textsc{Eq-YouCook2}}  &\multirow{2}{*}{\textsc{Eq-GEBC}}  &\multicolumn{3}{c}{\textsc{Eq-Kubric}}     &\multirow{2}{*}{\textsc{Eq-SD}}  &\multirow{2}{*}{Winoground}  & \multirow{2}{*}{Avg} \\ \cmidrule(lr){5-7}
&  &  &  & Location & Counting & Attribute &  & \\ \midrule
 FT (F30K) &9.24  &44.33  &8.54  & 8.95 & 28.49 & 66.54 &79.97  & 23.00  &33.63 \\
 + HardNeg &12.27  &45.43  &10.30  & 10.89 & 29.49 & 67.69 &81.69  & 27.00  &35.59  \\ 
 + \algnamev-all &12.15  &\textbf{45.71}  &\textbf{10.97}  & 9.79 & 29.94 & 68.75 &\textbf{81.78}  & 26.49  &35.69  \\ 
 + \algnamevv-all &12.10  &44.97  &10.08  & 10.25 & 29.05 & 68.30 &79.71  & 25.49  &34.99  \\ 
 + \algnamevv-close &\textbf{13.83}  &44.99  &10.80  & \textbf{11.15} & 29.25 & 69.25 &80.37  & 25.75  &35.67  \\  \hline
 + \algname &12.64  &45.10  &10.58  & 11.05   & \textbf{30.90} & \textbf{70.20}  &80.70  & \textbf{27.50}    &\textbf{36.08} \\ 
\shline
\end{tabular}
\caption{Full results of Table~\ref{tab:abla_arch}.}
\label{tab:supp_abla_arch}
\end{table*}

\subsection{Computation Cost of \algname}
We present the computation cost of adding \algname in the table below.
The forward time is measured with the average of 100 times of forward passes on a single GPU.
First, \algname is added as a regularization loss, \textbf{without} 
additional overhead on \# of parameters.
On time cost, we observe an acceptable overhead for fusion-encoder (\ie, METER) due to the similarity calculation on negative pairs.
While for dual-encoder (\ie, FIBER), which calculates the similarity for each image-text pair, the extra time needed for \algname is almost negligible.
Additionally, we show the forward time consumption \textit{v.s.} the batch size in 
the figure below.
The computation cost of \algname linearly scales with batch size, which is only slightly higher than the baseline for each data point.

  \begin{minipage}{0.95\linewidth}
  \vspace{1mm}
    \begin{minipage}[c]{0.42\linewidth}
    \centering
    \renewcommand{\arraystretch}{0.9}
        \scalebox{0.7}
        {\begin{tabular}{lcc}
        \shline
                   & \multicolumn{2}{c}{METER}  \\\cmidrule{2-3}
        Method     & Time  & Parameter  \\
        \midrule
        FT         & 0.384s         & 319M       \\
        \algname & 0.750s          & 319M         \\
        \midrule
                   & \multicolumn{2}{c}{FIBER}  \\\cmidrule{2-3}
        Method    & Time  & Parameter  \\
        \midrule
        FT         & 0.270s          & 242M      \\
        \algname & 0.277s         & 242M      \\
        \shline
        \end{tabular}}
      \label{tab:compare_cost}
    \end{minipage}
  \hfill
\begin{minipage}[c]{0.52\linewidth}
    \centering
    \footnotesize
    \vspace{-6mm}
    \includegraphics[width=1.0\linewidth]{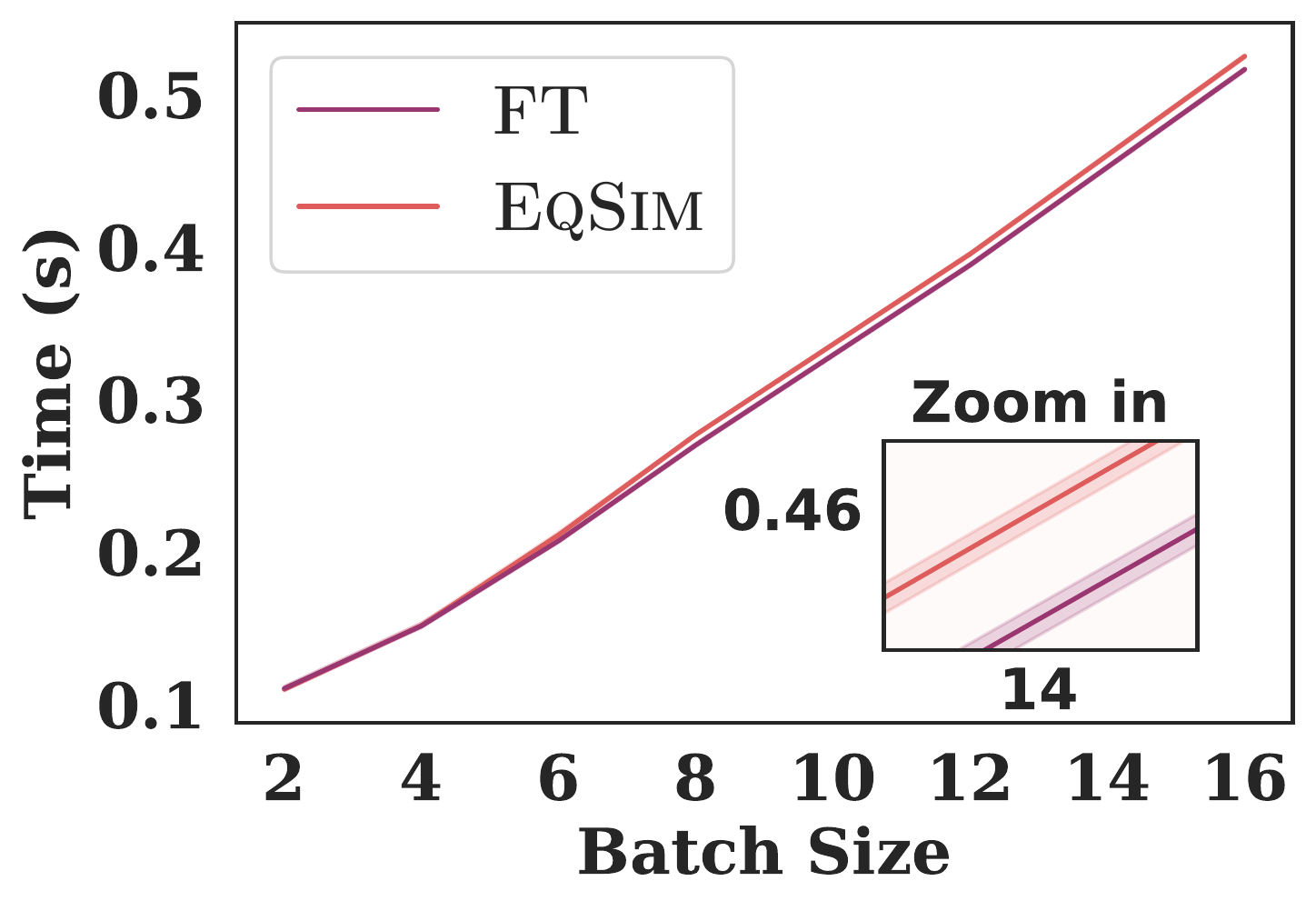}
    \vspace{-6mm}
    \label{fig:compare_cost}
    \vspace*{-4mm}
  \end{minipage}
  \end{minipage}

\subsection{More Benchmarking Results on \benchname}
\label{supp_sec:eqben_results}
We comprehensively report the model performance of existing VLMs on \benchname in Table~\ref{tab:supp_eqbench}. In addition to the observations drawn in the main paper, we can also find that: 1) When comparing the results of ALBEF/BLIP and their variants with contrastive loss (indicated by $\ddagger$), utilizing cosine similarity as the similarity measure as in ITC often leads to inferior accuracy compared to score computed by the ITM head. 
As ITC is usually implemented without cross-attention, making it hard to perform the fine-grained semantic recognition required in \benchname.
2) Fine-tuning on Flickr30K (F30K) results in a better performance. In contrast to the noisy samples of the pre-training data, F30K contains  high-quality captions that describe images in detail, hence helpful for the equivariant similarity learning of VLMs. 
3) The recent method BLIP2 shows strong capacity on our \benchname. Compared to other baselines, it is pre-trained on a much larger vision-language corpus (with 129 million image-text pairs), and thus shows better generalizability.
\begin{table*}[t]
\centering

\tablestyle{3pt}{1.2} 
\begin{tabular}{l|ccccccccc|cccccc|l}
\shline
\multirow{3}{*}{Method} & \multicolumn{9}{c|}{Natural Subsets}  & \multicolumn{6}{c|}{Synthetic Subsets}& \multirow{3}{*}{Avg}\\
\cmidrule(lr){2-10} \cmidrule(lr){11-16}
 & \multicolumn{3}{c}{\textsc{Eq-YouCook2}} & \multicolumn{3}{c}{\textsc{Eq-GEBC}} & \multicolumn{3}{c|}{\textsc{Eq-AG}} & \multicolumn{3}{c}{\textsc{Eq-Kubric}} & \multicolumn{3}{c|}{\textsc{Eq-SD}} &  \\
\cmidrule(lr){2-4} \cmidrule(lr){5-7} \cmidrule(lr){8-10} \cmidrule(lr){11-13} \cmidrule(lr){14-16}
   & Text & Image & Group & Text & Image & Group & Text & Image & Group & Text & Image & Group & Text & Image & Group &  \\ 
\hline
LXMERT~\cite{tan-bansal-2019-lxmert} &13.96  &11.98  &4.55  &13.56  &12.73  &4.19  &18.17  &9.02  &4.46  &18.50  &15.35  &7.26  &11.16  &6.15  &1.98   &10.20 \\
ViLBERT~\cite{lu2019vilbert} &14.78  &12.75  &5.18  &14.67  &12.64  &4.82  &17.43  &8.36  &3.89  &17.55 &18.44 &8.13  &12.37  &7.37  & 2.78  &10.74 \\
\hline
CLIP$^{\ddagger}$ (RN-50)~\cite{radford2021learning} & 47.72 & 47.99 & 34.05 & 10.80 & 18.03 & 3.97 & 14.52 & 10.44 & 3.50 & 21.33 & 21.93 & 9.75  &90.09 &  85.92 & 79.11 &  33.28\\
CLIP$^{\ddagger}$ (ViT-B/32)~\cite{radford2021learning} & 49.48 & 51.10 & 36.50 & 12.57 & 20.12 & 4.47 & 13.91 & 8.72 & 3.32 & 20.56 & 21.29 & 9.66 & 89.16 & 86.05 &78.98 &  33.73\\
FLAVA~\cite{singh2022flava} & 51.66 & 54.78 & 39.68 &12.24  &16.81  &5.07  & 6.59 & 13.47 & 2.15 & 28.88 & 28.18 & 15.90 &  79.64 & 84.47 & 71.10 &  34.04\\
ViLT~\cite{kim2021vilt} & 44.61 & 46.69 & 31.74 & 14.72 & 16.70 & 5.62 & 15.37 & 9.89 & 3.45 & 31.23 & 27.00 & 17.90 & 80.37 & 79.04 & 68.93 &  32.88\\
ViLT + FT (F30K) & 44.06 & 41.69 & 29.04 & 16.43 & 18.69 & 7.00 & 19.00 & 11.22 & 5.05  & 34.00 & 24.30 & 16.55 & 70.12 & 70.13 & 55.52 & 30.85 \\ \hdashline[2pt/5pt]
ALBEF$^{\ddagger}$~\cite{li2021align}  & 51.24 & 49.09 & 36.22 & 9.97 & 16.04 & 4.41 & 10.17 & 10.92 & 2.78  & 18.37 & 16.98 & 7.20 & 82.48 & 89.95 & 77.85 & 32.24 \\
ALBEF & 57.01 & 58.04 & 44.90 & 13.56 & 19.63 & 5.89 & 11.28 & 15.17 & 3.93 & 29.87 & 30.18  & 18.58 & 88.96 & 90.41 & 83.07 &  38.03\\
ALBEF$^{\ddagger}$ + FT (F30K) & 57.68 & 54.13 & 42.74 & 14.44 & 19.29 & 6.28 & 19.36 & 12.75 & 5.41 & 30.62 & 22.53 & 14.28 & 91.93 & 92.20 & 86.51 & 38.01 \\
ALBEF + FT (F30K) & 61.31 & 61.78 & 49.28 & 17.86 & 22.33 & 8.32 & 23.06 & 16.67 & 7.56 & 43.44 & 36.58 & 28.23 & 92.20 & 91.07 & 85.32 & 43.00 \\ \hdashline[2pt/5pt]
BLIP$^{\ddagger}$~\cite{li2022blip} & 55.48 & 55.60 & 42.42 & 13.84 & 18.68 & 6.39 & 13.02 & 10.46 & 3.72  & 27.03 & 26.03 & 14.88 & 82.28 & 85.26 & 74.48 & 35.30 \\
BLIP & 59.22 & 58.36 & 46.31 &  15.87 & 19.79 & 7.27 & 19.76 & 13.87 & 6.31 & 29.38 & 32.25 &  18.73 & 85.39 & 85.52 & 77.13  &  38.34 \\
BLIP$^{\ddagger}$ + FT (F30K) & 58.53 & 58.61 & 45.55 & 15.54 & 20.62 & 7.71 & 18.27 & 14.33 & 6.12  & 34.23 & 29.95 & 19.05 & 91.60 & 89.82 & 84.66 & 39.64 \\
BLIP + FT (F30K) & 65.30 & 64.00 & 52.96 & 20.45 & 23.65 & 10.08 & 23.02 & 18.97 & 8.03  & 46.10 & 38.56 & 28.93 & 92.39 & 92.20 & 86.31 & 44.73 \\ \hdashline[2pt/5pt]
BLIP2~\cite{li2023blip} &65.78 &69.48 &55.97 &18.25 &27.62 &9.15 &30.46 &20.39 &11.75 &40.24 &41.63 &28.53 &92.59 &91.54 &86.51 &45.19 \\
\hline
METER~\cite{dou2021empirical} & 52.18 & 49.42 & 36.81 & 20.95 & 18.19 & 6.95 & 28.70 & 15.80 & 7.88  & \textbf{44.28} & 35.20 & 27.26  & \textbf{89.62} & \textbf{84.93} & \textbf{79.44}  & 39.84\\
 + FT (F30K)~\cite{dou2021empirical} & 52.68 & 48.31 & 36.52 & 18.08 & 19.85 & 7.33 & \textbf{29.50} & 16.30 & 8.12 & 41.11 & 34.59 & 24.33  & 86.64 & 84.46 & 77.46  &  39.02\\
 + \algname &\cellcolor{mygray}\textbf{54.12}  &\cellcolor{mygray}\textbf{53.12}  &\cellcolor{mygray}\textbf{40.29} &\cellcolor{mygray}\textbf{24.20}  &\cellcolor{mygray}\textbf{26.02}  &\cellcolor{mygray}\textbf{11.69} &\cellcolor{mygray}28.85  &\cellcolor{mygray}\textbf{20.09}  &\cellcolor{mygray}\textbf{10.76} &\cellcolor{mygray}\underline{43.68}  &\cellcolor{mygray}\textbf{39.08}  &\cellcolor{mygray}\textbf{28.42}  &\cellcolor{mygray}\underline{88.04}&\cellcolor{mygray}84.07 &\cellcolor{mygray}\underline{77.79}  &\cellcolor{mygray}\textbf{42.28}  \\ 
\hline
FIBER~\cite{dou2022coarse} & 52.04 & 50.84 & 38.32 & \textbf{25.19} & 22.66 & \textbf{11.08} & \textbf{32.49} & \textbf{24.05} & \textbf{13.70} & 47.94 & 45.60 & 33.53 & 86.05 & \textbf{88.63} & 79.97  &  44.86\\
+ FT (F30K)~\cite{dou2022coarse} & 57.70 & 56.46 & 44.33 & 18.24 & 21.33 & 8.54 & 26.99 & 18.69 & 9.24 & 50.31 & 46.06 & 34.66 & 90.48 & 86.64 & \textbf{81.29}  &   43.40\\
 + \algname &\cellcolor{mygray}\textbf{58.26}  &\cellcolor{mygray}\textbf{57.10}  &\cellcolor{mygray}\textbf{45.10} &\cellcolor{mygray}\underline{21.55} &\cellcolor{mygray}\textbf{26.07} &\cellcolor{mygray}\underline{10.58} &\cellcolor{mygray}\underline{29.93}  &\cellcolor{mygray}\underline{23.42}  &\cellcolor{mygray}\underline{12.64}  &\cellcolor{mygray}\textbf{51.90}  & \cellcolor{mygray}\textbf{48.40} &\cellcolor{mygray}\textbf{37.38}  &\cellcolor{mygray}\textbf{90.81} & \cellcolor{mygray}85.98 &\cellcolor{mygray}80.70  &\cellcolor{mygray}\textbf{45.32} \\ 
\shline
\end{tabular}
\caption{Full results on \benchname. $\ddagger$ denotes using  cosine similarity of image and text representation as the similarity measure, following the common practice in ITC.
}
\label{tab:supp_eqbench}
\end{table*}

\subsection{Generalization to Video Grounding}
\label{supp_sec:generalization}

\begin{table*}[h!]
\centering
\tablestyle{6pt}{1.1} 
\begin{tabular}{llccccccccc}
\shline
\multicolumn{2}{c}{\multirow{2}{*}{Method}} & \multicolumn{8}{c}{Threshold (s)} & \multirow{2}{*}{Avg} \\
\multicolumn{2}{l}{} & 0.1 & 0.2 & 0.5 & 1.0 & 1.5 & 2.0 & 2.5 & 3.0 &  \\ \hline
\multirow{2}{*}{F30K} & FT & 2.27 & 5.18 & 13.33 & 26.04 & 36.16 & 45.21 & 53.15 & 60.01 & 30.16 \\
 & ~+~\algname & \cellcolor{mygray}\textbf{2.93} & \cellcolor{mygray}\textbf{6.11} & \cellcolor{mygray}\textbf{15.16} & \cellcolor{mygray}\textbf{27.06} & \cellcolor{mygray}\textbf{37.00} & \cellcolor{mygray}\textbf{46.14} & \cellcolor{mygray}\textbf{54.07} & \cellcolor{mygray}\textbf{60.76} & \cellcolor{mygray}\textbf{31.15} \\ \hline
\multirow{2}{*}{COCO} & FT & \textbf{2.89} & 5.81 & 14.27 & 26.56 & 36.52 & 45.25 & 53.13 & 60.19 & 30.57 \\
 & ~+~\algname & \cellcolor{mygray}2.82 & \cellcolor{mygray}\textbf{6.06} & \cellcolor{mygray}\textbf{15.22} & \cellcolor{mygray}\textbf{27.18} & \cellcolor{mygray}\textbf{36.85} & \cellcolor{mygray}\textbf{45.73} & \cellcolor{mygray}\textbf{53.68} & \cellcolor{mygray}\textbf{60.56} & \cellcolor{mygray}\textbf{31.01} \\ \hline
\multirow{2}{*}{F30K+COCO} & FT & 2.65 & 5.71 & 13.84 & 25.60 & 35.88 & 44.75 & 53.04 & 60.02 & 30.18 \\
 & ~+~\algname & \cellcolor{mygray}\textbf{3.04} & \cellcolor{mygray}\textbf{6.25} & \cellcolor{mygray}\textbf{14.89} & \cellcolor{mygray}\textbf{26.98} & \cellcolor{mygray}\textbf{36.73} & \cellcolor{mygray}\textbf{45.53} & \cellcolor{mygray}\textbf{53.2} & \cellcolor{mygray}\textbf{60.42} & \cellcolor{mygray}\textbf{30.88} \\ \shline
\end{tabular}
\caption{Comparison between our~\algname and baselines in video boundary grounding task with regard to different time thresholds.}
\label{tab:supp_generalization}
\end{table*}
To further validate the generalization ability of the proposed~\algname, we conduct additional experiments on a very different but relevant downstream task, zero-shot video boundary grounding task~\cite{wang2022geb+,shou2021generic}, where the model is required to accurately predict the video boundary indicating event status change, given the before and after query captions. To adapt a pre-trained VLM to this video-language task, we extract video frames at fps=5 first and measure the similarity between each frame and the two query captions. Then given the two adjacent frames ($I_1, I_2$) and the two query captions ($T_1, T_2$), we define a boundary grounding score $s_{bg}=s(I_1,T_1)+s(I_2,T_2)$ for boundary grounding, where $s$ is the similarity produced by VLMs. $s_{bg}$ actually measures whether the boundary is located between frame $I_1$ and $I_2$. The larger $s_{bg}$ means that $(I_1,T_1)$ and $(I_2,T_2)$ are more likely to be a simultaneous match, thus indicating the boundary between the before and after captions. 
Results are reported in Table~\ref{tab:supp_generalization} on metrics following ~\cite{wang2022geb+}. We compute the accuracies based on the absolute distance between ground truth time boundaries and the predicted time boundaries, with the threshold varying from 0.1s to 3s.
Across all compared baselines, our~\algname can attain consistent performance improvements on the average accuracy, suggesting that \algname is effective to identify fine-grained shot changes in videos.

\subsection{Distribution Curves on More Subsets}
\label{supp_sec:distribution_curve}
\begin{figure*}[t]
    \centering
    \includegraphics[width=.99\textwidth]{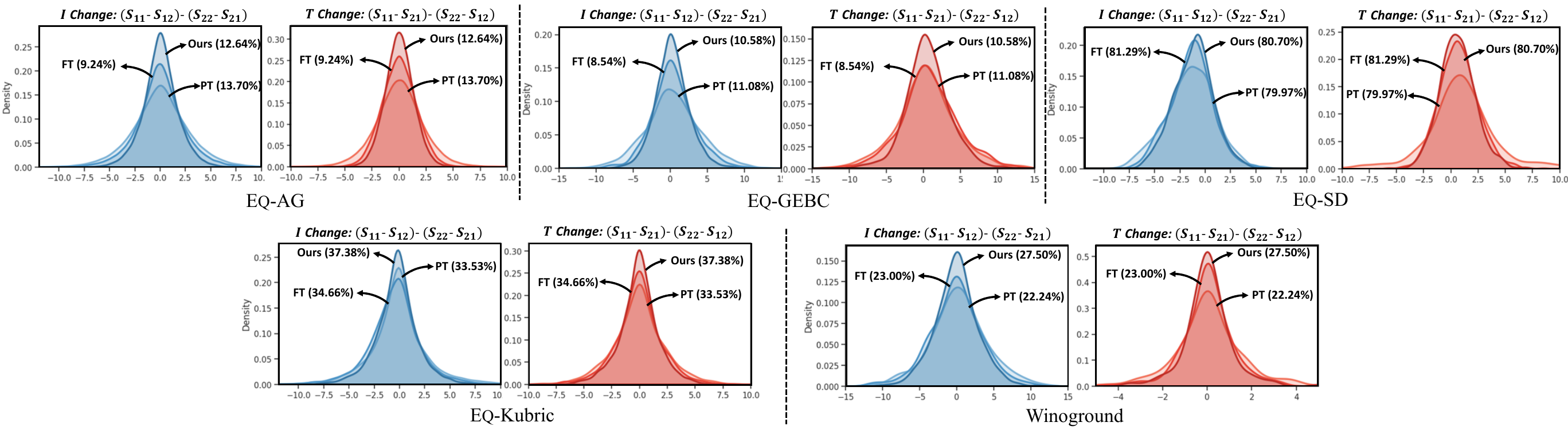}
    \caption{More visualizations of the equivariance score of baselines and our \algname based on FIBER~\cite{dou2022coarse} on other 4 subsets of \benchname.
    }
    \label{fig:sup_vis}
\end{figure*}

We present the distribution curves of the equivariant score on more \benchname subsets in Figure~\ref{fig:sup_vis} as the complement to Figure~\ref{fig:intro_hist} in the main paper. We can find that our \algname (indicated by ``Ours'') indeed achieves the most equivariant similarity (\ie, the tightest curve) across different datasets. Meanwhile, it is worth noting that the equivariance of similarity scores are not always positively correlated to the accuracy. 
For example, on \textsc{Eq-SD}, \algname (Ours) is similarly tight as the vanilla fine-tuning (FT), but the accuracy slightly drops.

\subsection{More Visualizations}
\label{supp_sec:visualization}
\noindent\textbf{Visualizations of Similarity Scores on Specific Examples.}
The distribution curves in Figure~\ref{fig:intro_hist} of the main paper depict the equivariant scores across the whole data. While in Figure~\ref{fig:sup_vis_eqben}, we explicitly visualize and compare the similarity scores (blue squares) for specific examples between FIBER baseline and our \algname.
We can clearly observe that current SoTA VLM still falls short in the similarity measure when facing two visually similar images. On one hand, the matching results are not even correct (red cross); On the other hand, regardless of the correctness of the matching results, the similarity scores are not yet equivariant, similar to the Figure~\ref{fig:intro_fig}(b) of the main paper. As shown in the left part of Figure~\ref{fig:sup_vis_eqben}, for the same visual semantic change (\texttt{red}$\leftrightarrow$\texttt{grey}), the corresponding similarity change of FIBER $s_{11}-s_{21}=-0.07$ is very different from $s_{22}-s_{12}=1.4$. While our model can produce much more equivariant similarity measure.

\begin{figure}[h!]
    \centering
    \vspace{-0.1in}
    \includegraphics[width=.48\textwidth]{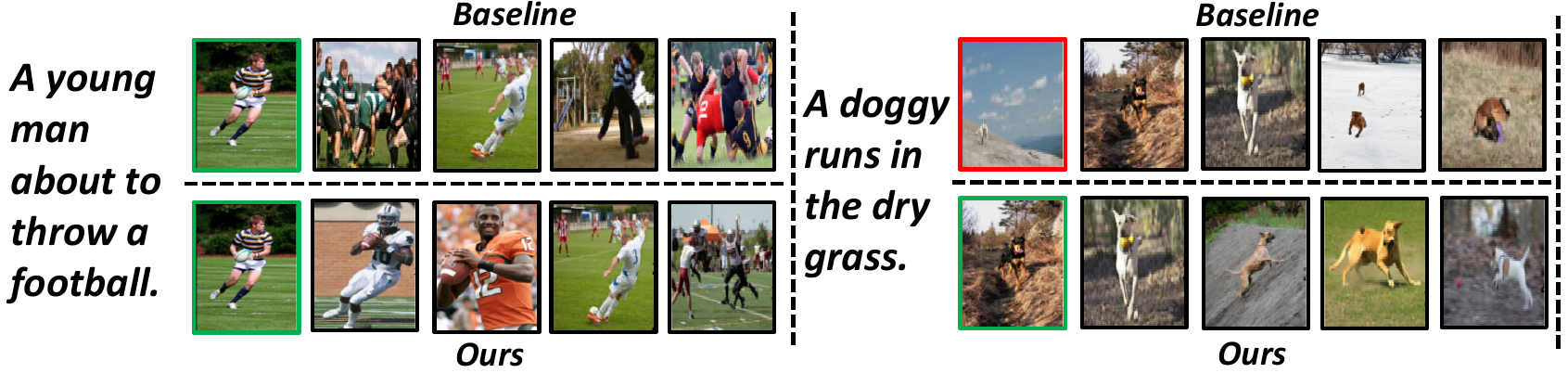}
    \caption{Visualization of top-$5$ T2I retrieval results on Flickr30K. Correct (wrong) top-1 images are in green (red).}
    \label{fig:vis_f30k}
\end{figure}

\noindent\textbf{Visualizations of Retrieval Results.}
We visualize the retrieval results on the commonly used Flickr30K in Figure~\ref{fig:vis_f30k}. In addition to the better top-$1$ retrieval accuracy, our \algname can produce much more reasonable similarity measurements for the whole retrieval sequence. For example, for the baseline model, the rank $2$ and $3$ images are \textbf{not} in line with the text of ``a young man'' and `` throw''. While our top-$3$ images are clearly more relevant.

\subsection{\algname vs. CyCLIP~\cite{goel2022cyclip}}

 We notice this related contemporaneous work~\cite{goel2022cyclip}. We compare and discuss more detailed differences here.
\begin{itemize}
    \item Different motivation and implementation. Given the two image-text pairs $\{I_1,T_1\}$ and $\{I_2,T_2\}$, CyCLIP regularizes the CLIP cosine similarity score $s$ with the in-modal consistency (forcing $s(I_1,I_2)$ to be close to $s(T_1,T_2)$) and the cross-modal consistency (forcing $s(I_1,T_2)$ to be close to $s(I_2,T_1)$). While our \algname steps from the motivation that the similarity score change should faithfully respect to the semantic change and derive to the two regularization terms in Eq.~\eqref{eq:method_eq6}. Our final objective is the weighted combination of such two terms.
    \item Different evaluation settings and tasks. CyCLIP is solely built on dual-encoder architecture (\eg CLIP) and evaluates the effectiveness on the zero-shot image classification task. While our~\algname can adapt to both dual-encoder and fusion-encoder architectures (\eg METER and FIBER) and achieve improvements across various  VL benchmarks and downstream tasks, \eg, image-text retrieval, vision-language compositionality, and video boundary grounding.
    \item Better performance of~\algname. The closest CyCLIP counterpart to our~\algname  is the cross-modal consistency, which we implemented as \algnamev-all in Table~\ref{tab:abla_arch}. As we compare \algnamev-all against  +\algname, we clearly observe the superior performance of our \algname.
\end{itemize}

\begin{figure}[h!]
    \centering
    \includegraphics[width=.48\textwidth]{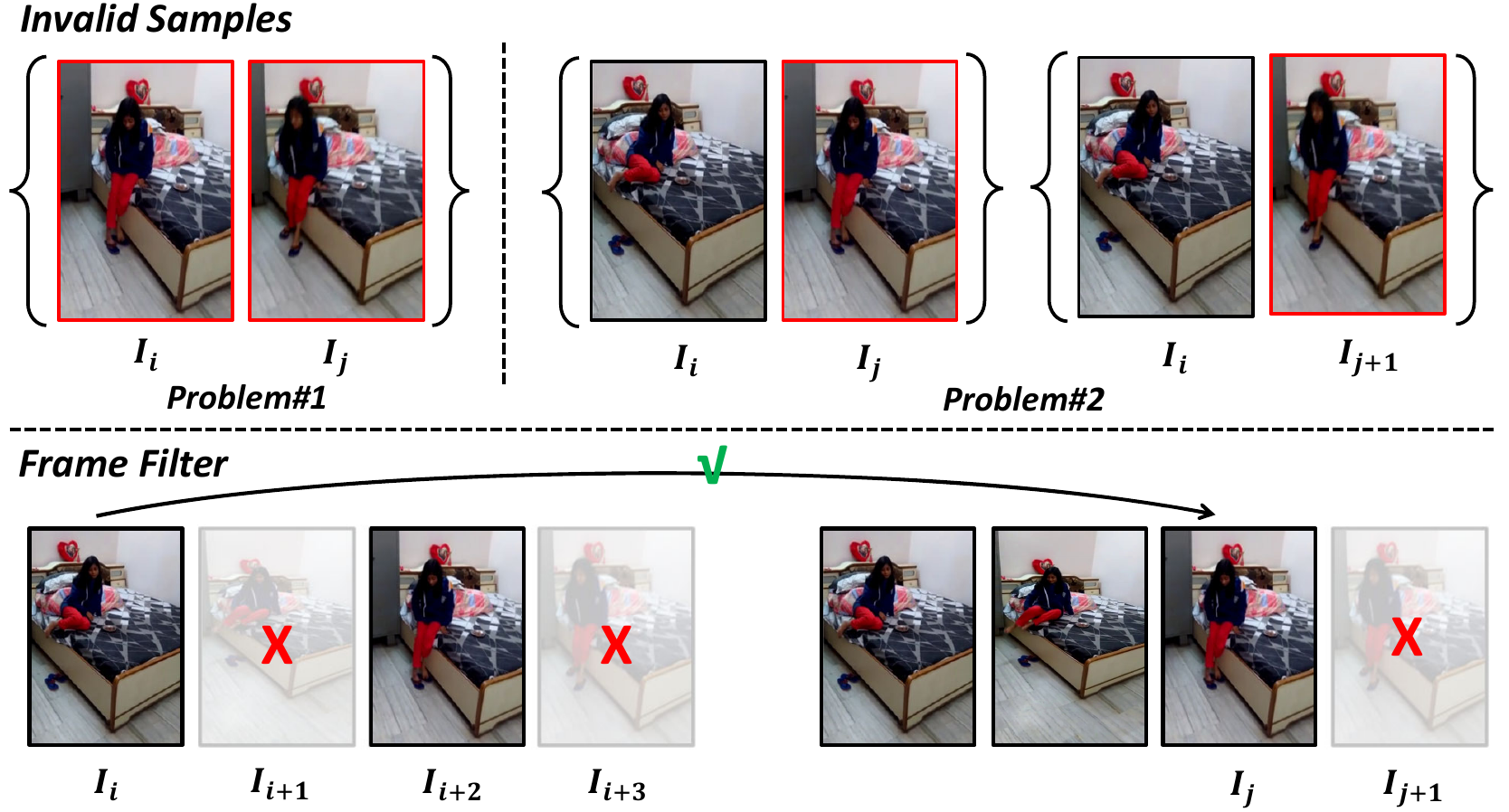}
    \caption{The invalid examples for AG (top) and our proposed frame filter (bottom).
    }
    \label{fig:sup_eqag}
\end{figure}

\section{Construction Details of \benchname}
\label{sec:data}

In addition to the general construction pipeline of \benchname in Section~\ref{sec:exp_eqbench} of the main paper, here we include more details specific to each subset.
For all subsets built on natural videos, we denote $I_i$ and $I_j$ as two different frames of the video, while $I_{i+1}$ ($I_{j+1}$) represents the immediate next frame following $I_i$ ($I_j$).
\subsection{\textsc{Eq-AG}}

\noindent\textbf{Source Dataset}.
Action Genome~\cite{ji2020action} (AG) captures changes between objects and their pairwise relationships while action occurs. It contains nearly 10K videos with 1.7M visual relationships which can be used for caption generation. Given the scene graph $\langle$person - \textit{attention relationship} - \textit{spatial relationship} - object$\rangle$, we first create the caption with the template ``The person is $\langle$\textit{attention relationship}$\rangle$ $\langle$\textit{object}$\rangle$ which is $\langle$\textit{spatial relationship}$\rangle$ him/her.''

\noindent\textbf{Invalid Samples}. In AG, we find that sometimes it is hard to tell apart the two adjacent frames due to the continuity of the video data. This results in two problems for the dataset construction as shown in Figure~\ref{fig:sup_eqag} (top): 1) The two images $I_i$ and $I_j$ are too similar, and may be described by the same caption; 2) The two sample pairs $\{I_i, I_j\}$ and $\{I_i, I_{j+1}\}$ are too similar, leading to many duplicates. 

\begin{figure}[h!]
    \centering
    \includegraphics[width=.48\textwidth]{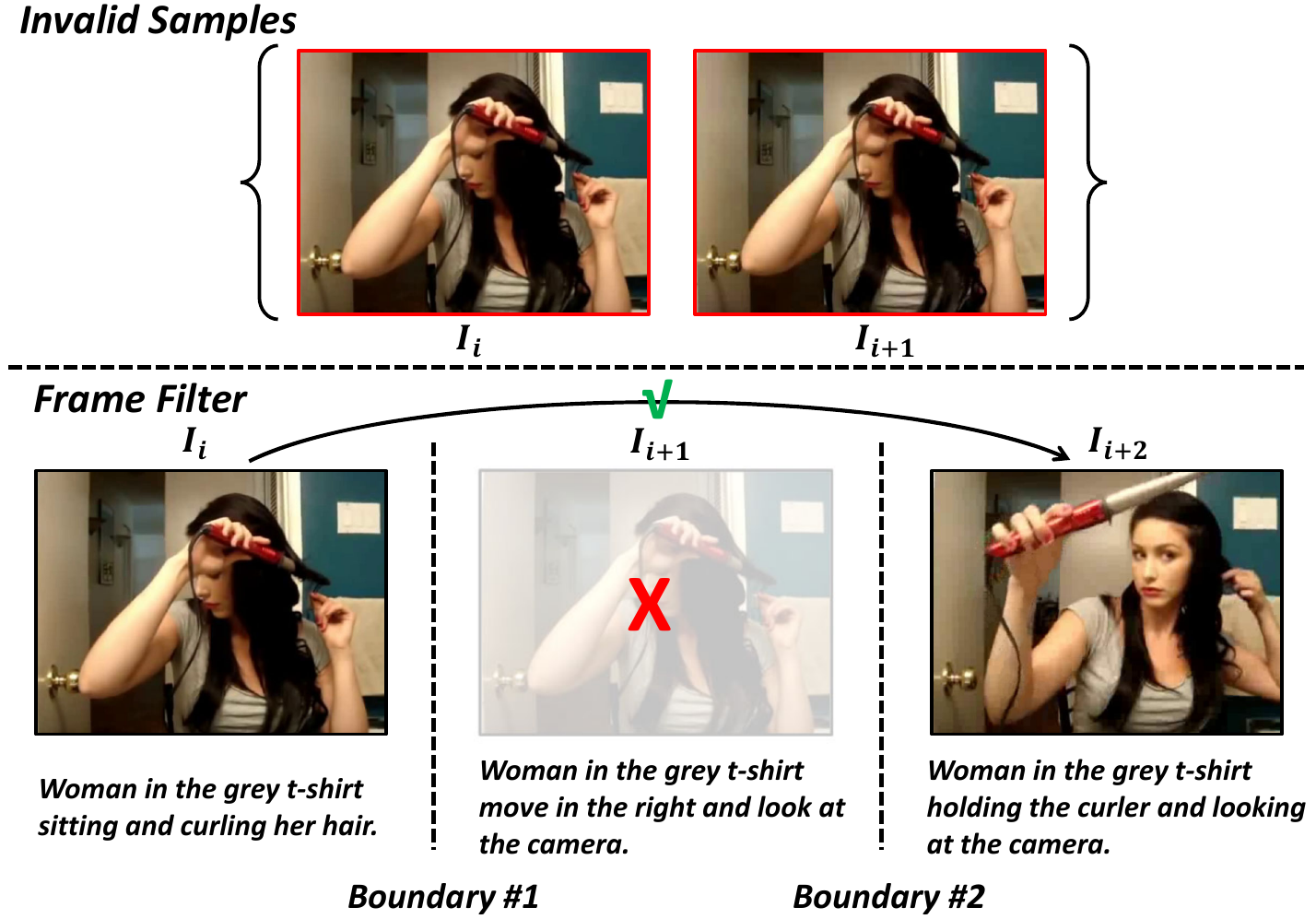}
    \caption{The invalid samples for GEBC (top) and our proposed frame filter (bottom). 
    }
    \label{fig:sup_eqgebc}
\end{figure}

\noindent\textbf{Frame Filter}.
To solve this problem, we adopt a sparse sampling strategy to select frames as candidates (Figure~\ref{fig:sup_eqag} (bottom)). Specifically, we only choose frames $I_i$ and $I_j$ if and only if at least 2 of 3 relationships are different. This will make sure the distinction between two images in a single sample, thus solving problem \#1.
Furthermore, for problem \#2, we assume that given a chosen frame $I_i$ ($I_j$), the immediate next frame $I_{i+1}$ ($I_{j+1}$) is too similar to $I_i$ ($I_j$). Therefore, if $I_i$ ($I_j$) is chosen, we will skip the subsequent frame (red cross in Figure~\ref{fig:sup_eqag} (bottom)), and move to $I_{i+2}$ ($I_{j+2}$). 

\begin{figure}[h!]
    \centering
    \includegraphics[width=.48\textwidth]{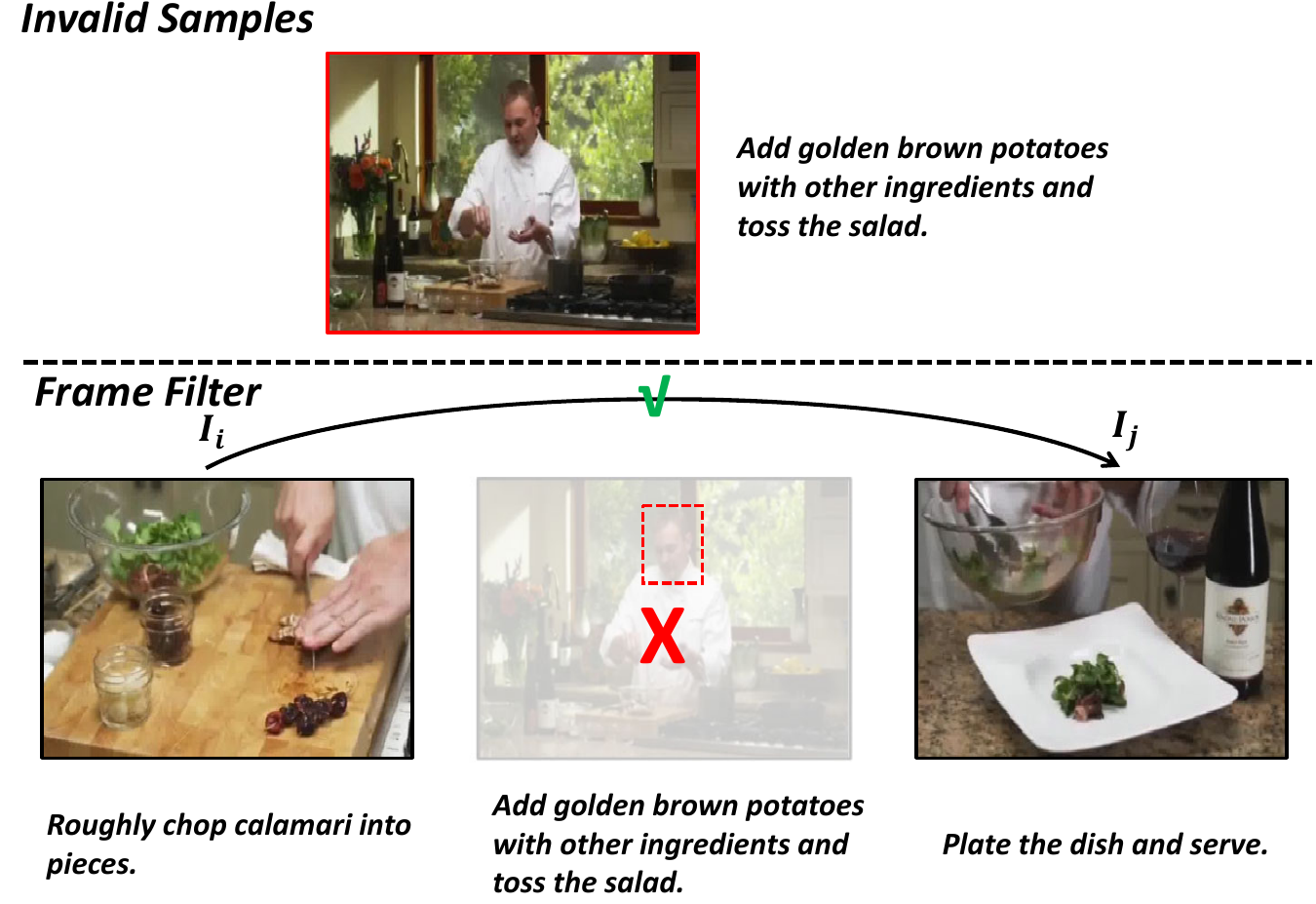}
    \caption{The invalid samples for YouCook2 (top) and our proposed frame filter (bottom). 
    }
    \label{fig:sup_eqyoucook}
\end{figure}

\subsection{\textsc{Eq-GEBC}}
GEBC~\cite{wang2022geb+} consists of over 170k boundaries associated with captions describing the events before and after the boundaries. It is built upon 
 12K videos from Kinetic-400~\cite{kay2017kinetics} dataset. We construct \textsc{Eq-GEBC} examples based on annotations from the training and validation splits of GEBC.
Intuitively, 
we can directly adopt the frames before and after the boundaries (\ie, $I_i$ and $I_{i+1}$) as our visual minimally different images, and the provided GEBC annotation before and after the boundaries can be naturally leveraged as the captions.

\noindent\textbf{Invalid Samples}. However, similar to AG, we find that it is hard to tell apart the two images separated by a boundary in practice (see Figure~\ref{fig:sup_eqgebc} (top)). The reason behind is that the boundary of GEBC is annotated as the status change between two video segments (\eg, from ``walking'' to ``running''). Such action words can be hard to recognize from the sampled static frames.

\begin{figure*}[t]
    \centering
    \includegraphics[width=.95\textwidth]{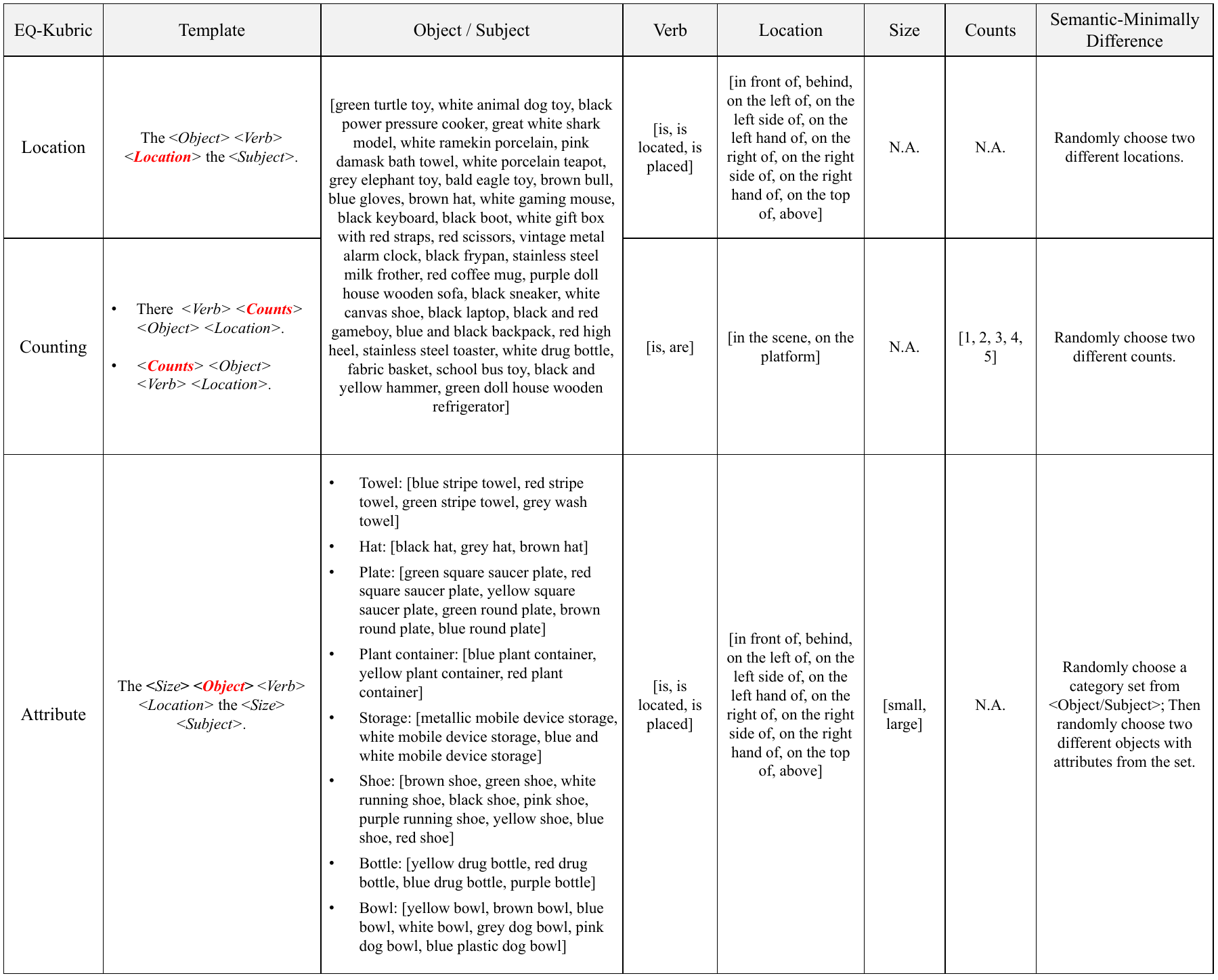}
    \caption{Overview of caption generation pipeline for three subsets (\ie, location, counting and attribute) of  \textsc{Eq-Kubric}. Text in red indicates the aspect of semantic change between two captions.
    }
    \label{fig:sup_eqkubric}
\end{figure*}

\noindent\textbf{Frame Filter}.
As shown in Figure~\ref{fig:sup_eqgebc} (bottom), we propose to skip an additional boundary to choose $I_i$ and $I_{i+2}$ as the twin images to enlarge the semantic gap. Meanwhile, we filter out images with captions containing action words (\eg, ``up'', ``down'', ``upward'', ``downward'' and ``towards''), which are hard to infer without temporal information. Finally, to ensure data quality, we perform a manual screening process with 10 graduate students to filter out invalid samples.

\begin{figure*}[t]
    \centering
    \includegraphics[width=.95\textwidth]{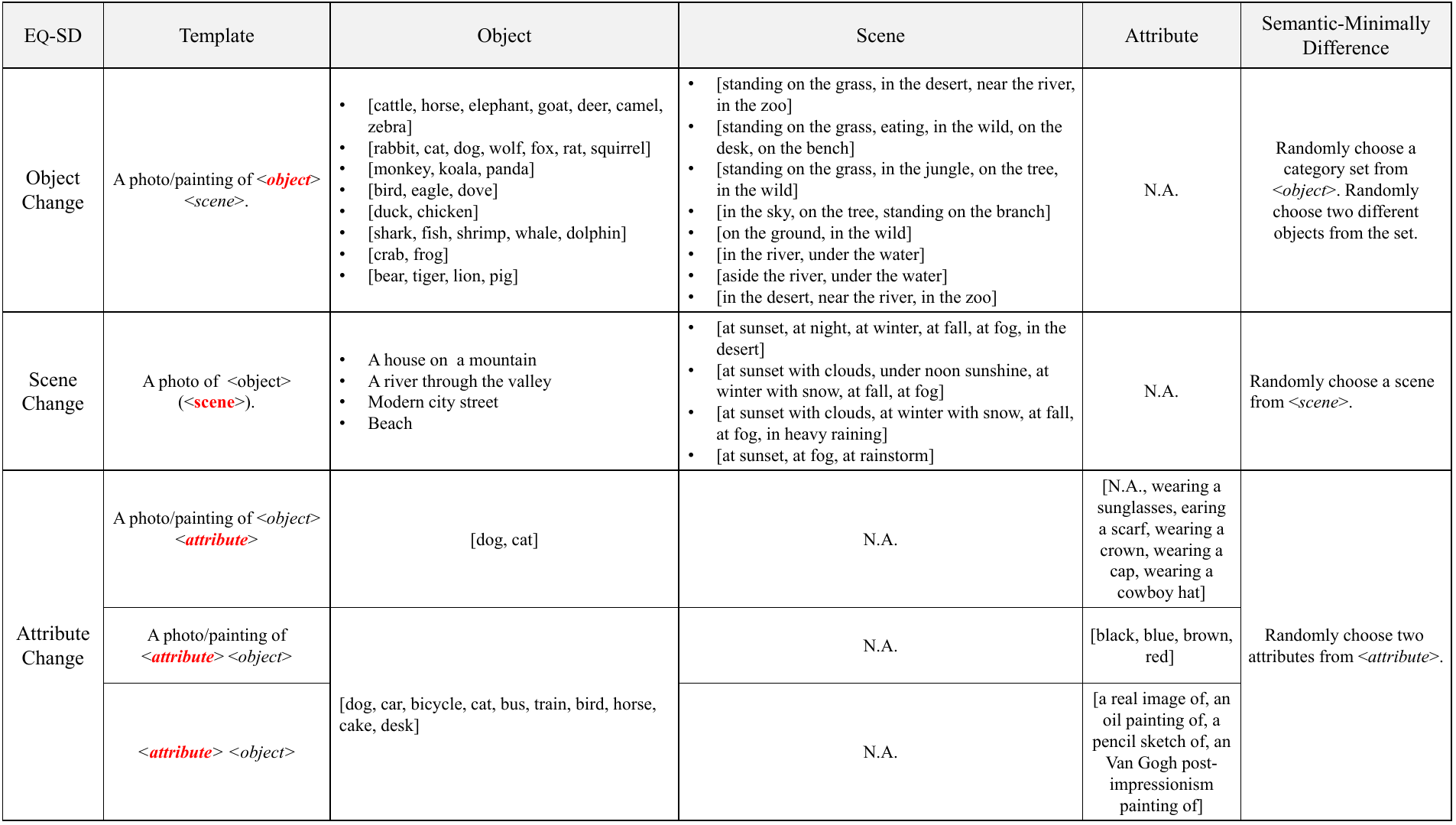}
    \caption{Overview of caption generation for  \textsc{Eq-SD} dataset. The red color highlights the aspect of semantic change between two captions. 
    }
    \label{fig:sup_eqsd}
\end{figure*}

\noindent\textbf{Invalid Samples}. As shown in Figure~\ref{fig:sup_eqyoucook} (top), we find that for the cooking video, the chosen frame may contain the view of the chef rather than accurately capturing the objects described in the cooking step. This leads to the mismatch between the image and the caption.

\subsection{\textsc{Eq-YouCook2}}
We utilize YouCook2~\cite{zhou2018towards} as the data source which contains 2K YouTube videos with average duration of 5.3 minutes, summing to a total of 176 hours. The videos have been manually annotated with segmentation boundaries and captions. 
On average there are 7.7 segments/captions per video, and 8.8 words per caption. 
 We construct \textsc{Eq-YouCook2} examples based on annotations from the training and validation splits of YouCook2.
For each video with $N$ segments, we directly select the middle frame as $I_i$ and its annotated caption as $T_i$, $i \in \{1, 2, ..., N\}$.

\noindent\textbf{Frame Filter}.
To solve this problem, we adopt a simple yet effective solution with the face detector~\footnote{\url{https://github.com/ageitgey/face_recognition}} for frame filtering. Specifically, we directly discard the frames with human faces.

\subsection{\textsc{Eq-Kubric}}
As introduced in the main paper, \textsc{Eq-Kubric} takes advantage of an open-source graphics engine~\cite{greff2021kubric} to faithfully generate photo-realistic scene for the given captions. Therefore, the visual-minimal images generation has been translated into the semantic-minimally different captions construction. We categorize the caption change into three aspects: \textit{attribute}, \textit{counting} and \textit{location}. Figure~\ref{fig:sup_eqkubric} presents the caption construction details. The semantic-minimally difference is ensured by only intervening the corresponding part in the template while leaving other words unchanged.

\subsection{\textsc{Eq-SD}}
Similar process can be applied to \textsc{Eq-SD}, for which we summarize the construction details in Figure~\ref{fig:sup_eqsd}. We similarly categorize the textual semantic-minimal editing into three aspects: object change, scene change and attribute change. We randomly select from the aforementioned three aspects to construct the semantic-minimally different captions.
However, in contrast to the Kubric engine, the generation quality of the stable diffusion model is heavily correlated to the given textual prompt. Therefore, we design a more fine-grained template selection for SD.  We select $\langle$\textit{scene}$\rangle$ and $\langle$\textit{attribute}$\rangle$ from a more restricted subset based on $\langle$\textit{object}$\rangle$. For example, given the object of ``horse'' and the category of ``object change'' (first row of Figure~\ref{fig:sup_eqsd}), the changed object will be selected from the animals from the same subset (\ie, ``cattle'', ``elephant'', ``goat'', ``deer'', ``camel'' and ``zebra''). Meanwhile, the scene shared across two captions will be selected from the first subset of scene (\ie, ``standing on the grass'', ``in the desert'', ``near the river'' and ``in the zoo'') for rationality. 

\begin{figure*}[t]
    \centering
    \includegraphics[width=.99\textwidth]{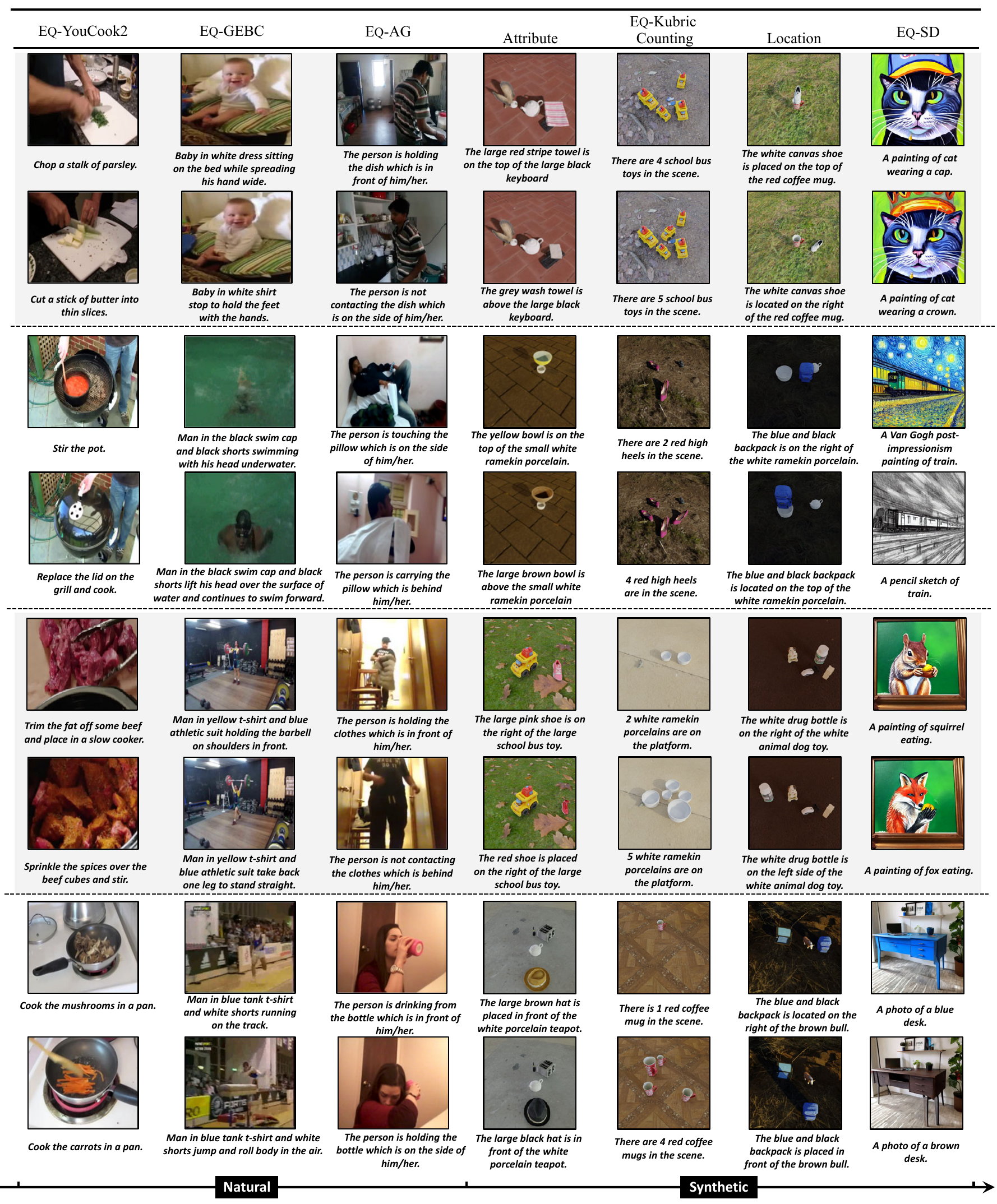}
    \caption{More visualizations of the examples for our \benchname.
    }
    \vspace{10pt}
    \label{fig:sup_example}
\end{figure*}

\section{Implementation Details of \algname}
\label{sec:implement}

We fine-tune the models on 8 NVIDIA V100 GPUs.
The regularization margin $\alpha$ and the balancing factor $\beta$ are selected from $\{0, 0.04, 0.1\}$ and $\{0.2, 0.5, 1.0\}$. We adopt an image resolution as $288\times288$ due to computational constraints. 
For FIBER, which is implemented with ITC loss for fast retrieval, 
we adopt the cosine similarity between image and text features as $s$, and then normalize it by a softmax function. The images and text with top-$8$ $s$ are regarded as the semantically ``close'' samples to apply \algnamevv.
METER is designed with ITM loss, which does not compute all pairwise similarities in the training batch.  
Therefore, we leverage the pre-trained METER model to pre-compute and cache all pairwise similarities in Flickr30K training split, prior to fine-tuning. However, the computation of the ITM similarity for each image-text pair of the training set still takes a long time (more than two weeks on 8 V100 GPUs in practice). To further reduce the computation, we apply a ``coarse-to-fine'' strategy. For a given image, we first select the top-128 similar images, based on the image feature extracted from the METER vision encoder. Assuming each image is associated with 5 captions, we then utilize the ITM head to compute a fine-grained similarity measure for $128\times 5$ image-text pairs (leading to 1 hour on 8 V100 GPUs). 
During retrieval fine-tuning, we follow the original METER to sample $15$ captions as negatives and additionally sample their counterpart images for \algname. Furthermore, $8$ of $15$ items (\ie, $k=8$) are selected as hard negative (\ie, semantically close) samples based on the pre-computed similarity matrix. 
In METER, the similarity score $s$ is normalized with a sigmoid activation. 
We apply other model-specific hyper-parameters (\eg, training epochs and learning rates) following the original METER~\cite{dou2021empirical} and FIBER paper~\cite{dou2022coarse}.

\section{More Examples of \benchname}
\label{sec:example}

Figure~\ref{fig:sup_example} visualizes examples in \benchname. We can clearly find that the two images from one data sample  are visually similar, indicating that our \benchname indeed focuses on visual-minimal change.


\clearpage
{\small
\bibliographystyle{ieee_fullname}
\bibliography{egbib}
}

\end{document}